\documentclass{article}
\usepackage{iclr2027_conference,times}
\iclrfinalcopy

\usepackage{amsmath,amsfonts,bm}

\def\eqref#1{equation~\ref{#1}}

\def\1{\bm{1}}

\DeclareMathAlphabet{\mathsfit}{\encodingdefault}{\sfdefault}{m}{sl}
\SetMathAlphabet{\mathsfit}{bold}{\encodingdefault}{\sfdefault}{bx}{n}

\usepackage{hyperref}
\usepackage{url}
\usepackage{amsmath,amssymb}
\usepackage{booktabs}
\usepackage{graphicx}
\usepackage{xcolor}
\usepackage{tabularx}
\usepackage{array}
\usepackage{multirow}
\usepackage{microtype}
\usepackage{placeins}
\usepackage{float}
\usepackage{listings}
\usepackage[most]{tcolorbox}
\definecolor{systembg}{HTML}{F3EFF9}
\definecolor{skillbg}{HTML}{F6F2FB}
\definecolor{userbg}{HTML}{F7F7F7}
\definecolor{policybg}{HTML}{F1F5FC}
\definecolor{observationbg}{HTML}{F2F8EF}
\definecolor{feedbackbg}{HTML}{FFF8E8}
\definecolor{creditbg}{HTML}{EEF8F7}
\definecolor{neutralbg}{HTML}{F6F6F6}
\definecolor{trajectoryframe}{HTML}{6B4DA3}

\newtcolorbox{trajectorybox}[1]{
    enhanced jigsaw,
    breakable,
    colback=white,
    colframe=trajectoryframe,
    colbacktitle=systembg,
    coltitle=black,
    fonttitle=\small\bfseries,
    title={#1},
    title after break={#1 (continued)},
    boxrule=0.75pt,
    arc=2pt,
    left=5pt,
    right=5pt,
    top=5pt,
    bottom=5pt,
    before skip=7pt,
    after skip=9pt
}

\newtcolorbox{tracebox}[2][neutralbg]{
    enhanced,
    breakable,
    width=\linewidth,
    colback=#1,
    colframe=#1!65!black,
    colbacktitle=#1,
    coltitle=black,
    fonttitle=\small\bfseries,
    fontupper=\small,
    title={#2},
    title after break={#2 (continued)},
    boxrule=0.45pt,
    arc=1.5pt,
    left=5pt,
    right=5pt,
    top=4pt,
    bottom=4pt,
    before skip=4pt,
    after skip=4pt
}

\lstdefinestyle{appendixbase}{
    basicstyle=\ttfamily\footnotesize,
    frame=single,
    rulecolor=\color{gray!35},
    framerule=0.35pt,
    framesep=4pt,
    framexleftmargin=3pt,
    framexrightmargin=3pt,
    breaklines=true,
    breakatwhitespace=false,
    columns=fullflexible,
    keepspaces=true,
    showstringspaces=false,
    aboveskip=0.5\baselineskip,
    belowskip=0.5\baselineskip
}
\lstdefinestyle{appendixcode}{style=appendixbase,backgroundcolor=\color{neutralbg}}
\lstdefinestyle{skillcode}{style=appendixbase,backgroundcolor=\color{skillbg}}
\lstdefinestyle{policycode}{style=appendixbase,backgroundcolor=\color{policybg}}
\lstdefinestyle{observationcode}{style=appendixbase,backgroundcolor=\color{observationbg}}
\lstdefinestyle{feedbackcode}{style=appendixbase,backgroundcolor=\color{feedbackbg}}
\lstdefinestyle{creditcode}{style=appendixbase,backgroundcolor=\color{creditbg}}
\lstdefinestyle{tracecontent}{
    basicstyle=\ttfamily\footnotesize,
    frame=none,
    breaklines=true,
    breakatwhitespace=false,
    columns=fullflexible,
    keepspaces=true,
    showstringspaces=false,
    aboveskip=0pt,
    belowskip=0pt
}
\newtcblisting{tracelisting}[2][neutralbg]{
    enhanced,
    breakable,
    width=\linewidth,
    colback=#1,
    colframe=#1!65!black,
    colbacktitle=#1,
    coltitle=black,
    fonttitle=\small\bfseries,
    title={#2},
    title after break={#2 (continued)},
    boxrule=0.45pt,
    arc=1.5pt,
    left=5pt,
    right=5pt,
    top=4pt,
    bottom=4pt,
    before skip=4pt,
    after skip=4pt,
    listing only,
    listing options={style=tracecontent}
}

\newcommand{\bluecite}[1]{\textcolor{blue}{\citep{#1}}}

\newcommand{\code}{\texttt{<code>...</code>}}

\title{Beyond Instruction Following: Learning Grounded Skill-Following with Skill Contracts}

\author{%
\begin{minipage}{\dimexpr\textwidth-2\tabcolsep\relax}
\centering\footnotesize
Jianghan Shen\textsuperscript{1,2}, Zhenjie Liu\textsuperscript{1},
Yue Li\textsuperscript{3,4}, Jie Huang\textsuperscript{2},
Siqi Luo\textsuperscript{2}, Yiming Cheng\textsuperscript{5},
Yizhi Yao\textsuperscript{4}\\
Kaijie Zhang\textsuperscript{6},
Cheng Tang\textsuperscript{2}, Minghui Zhang\textsuperscript{2},
Ming Hu\textsuperscript{2}, Yirong Chen\textsuperscript{2},
Ziyan Huang\textsuperscript{2,*}\\[0.7em]
\normalfont\footnotesize
\textsuperscript{1}Nanjing University\quad
\textsuperscript{2}Shanghai Artificial Intelligence Laboratory\quad
\textsuperscript{3}Peking University\\
\textsuperscript{4}Yiyue Technology\quad
\textsuperscript{5}Tsinghua University\quad
\textsuperscript{6}Northeastern University
\end{minipage}%
}

\begin{document}
\raggedbottom
\maketitle
\begingroup
\renewcommand{\thefootnote}{\fnsymbol{footnote}}
\footnotetext[1]{Corresponding author: Ziyan Huang. Email: \href{mailto:huangziyan@pjlab.org.cn}{huangziyan@pjlab.org.cn}.}
\endgroup
\lhead{}

\begin{abstract}
Instruction following typically enforces discrete, response-level requirements, whereas an expert-authored skill prescribes procedural requirements spanning multiple phases and environment interactions.\ %
Given such a skill, we aim to train the executor to execute all required phases successfully instead of focusing solely on the final answer.\ %
We therefore introduce \textbf{Grounded Skill-Following}, which requires an agent to execute a fixed, expert-authored skill across its required phases by grounding decisions in environment observations.\ %
To achieve verifiable procedural execution, we formulate each skill as a \emph{skill contract} combining visible skill instructions with an explicit \emph{contract runtime}.\ %
The contract runtime specifies the required phases, admissible actions, permitted transitions, and accepted termination.\ %
This contractual structure naturally provides a dense, verifiable training signal throughout execution.\ %
We leverage this by introducing \emph{Verified Progress Credit}, which assigns rewards upon the initial completion of contract milestones and aggregates them into the trajectory return to guide policy optimization.\ %
During rollout, the contract runtime continuously tracks state transitions to provide \emph{Contract-State Feedback}, which indicates whether the latest action is accepted and guides the agent toward valid next actions.\ %
To measure procedural compliance, we introduce the \textbf{Protocol Completion Rate (PCR)}---defined as reaching accepted termination through all required phases---and decouple it from the final Task Outcome.\ %
Jointly trained with our framework, Qwen3.5-4B achieves Protocol Completion Rates of 99.27\% on Math and 99.96\% on Search, while slightly outperforming original baselines in Task Outcome (82.95\% and 46.61\%, respectively).\ %
Controlled studies examine how skill instructions, training signals, and contract-state feedback affect both metrics, while withholding interventions evaluate behavioral dependence on observation content.\ %
Project repository: \url{https://github.com/MrToser/skill-following}.
\end{abstract}

\section{Introduction}

\begin{figure}[!tbp]
\centering
\vspace{-4mm}
\includegraphics[width=\textwidth]{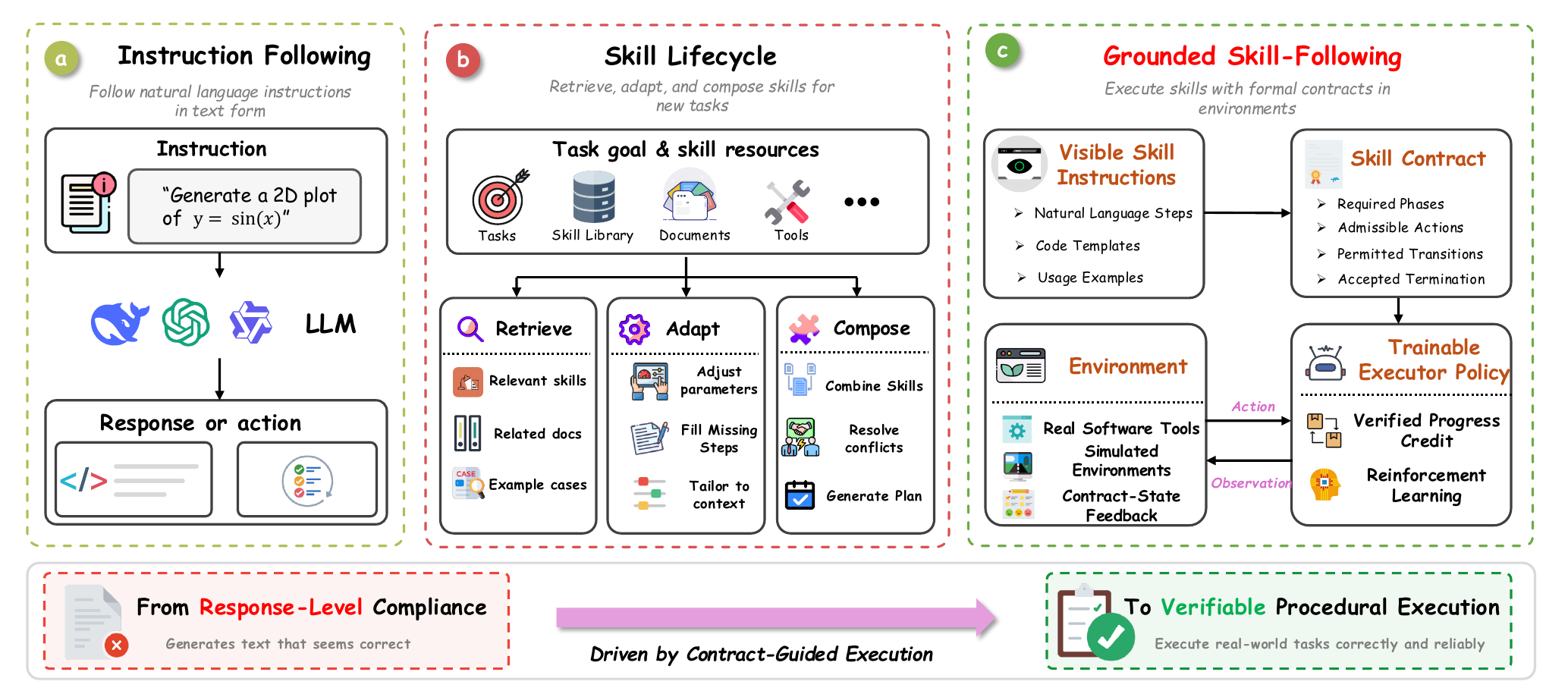}
\vspace{-8mm}
\caption{Complementary views of skills. Grounded skill-following fixes an expert-authored skill and trains its executor, alongside local instruction compliance and skill-lifecycle optimization. The views overlap and are not a historical taxonomy.}
\vspace{-4mm}
\label{fig:skill-perspectives}
\end{figure}

LLMs are trained to generate responses that satisfy natural-language instructions, a capability commonly referred to as \emph{instruction following}~\bluecite{wei2022finetuned,sanh2022multitask,ouyang2022training}, with existing benchmarks evaluating compliance mainly on discrete, response-level requirements~\bluecite{zhou2023ifeval,wen2024complexbench}(Figure~\ref{fig:skill-perspectives}(a)).
Tool-using agents extend this capability by grounding actions and decisions in environment observations~\bluecite{schick2023toolformer,nakano2021webgpt,yao2023react}.
Recent systems package instructions, tools, and execution strategies into reusable skills~\bluecite{anthropic2025skills,wang2023voyager,wang2024awm}.
Here, an expert-authored \emph{skill} prescribes procedural requirements spanning multiple phases and environment interactions, defining the order of admissible actions, the reliance on observation evidence, and valid completion conditions.
Research on the skill lifecycle spans retrieval~\bluecite{li2025sage,su2026skillretrieval}, adaptation~\bluecite{liu2026skiller,chen2026skill0}, composition~\bluecite{yang2026skillrl,zhang2026skill1,zhang2026skillcomposer}, and verified execution~\bluecite{skillcoach2026,tan2026skt} (Figure~\ref{fig:skill-perspectives}(b)).
While these efforts effectively acquire, adapt, or evaluate skills, they rarely enforce or evaluate whether an agent strictly adheres to prescribed procedural constraints across all required phases.
Because an agent can produce a correct final answer while bypassing the skill's required procedure, outcome correctness alone is insufficient to establish procedural compliance.

We term this unaddressed setup \textbf{Grounded Skill-Following}, a setting in which an expert-authored skill remains fixed and visible, and the executor must strictly fulfill all required phases (Figure~\ref{fig:skill-perspectives}(c)).
Here, following the prescribed procedure is itself part of the task requirements.
Methods such as Search-R1~\bluecite{jin2025searchr1} and ReTool~\bluecite{deepseekai2025retool} optimize executors through reinforcement learning, primarily targeting final Task Outcomes.
These outcomes alone do not establish procedural compliance.
Moreover, when rewards require completing the entire procedure, unsuccessful trajectories can receive identical rewards despite making different progress.
The challenge is therefore to make skill requirements verifiable during execution and provide learning signals that distinguish intermediate procedural progress.

To operationalize fixed skill specifications during agent-environment interaction, we represent each skill as a \emph{skill contract}.
Beyond static prompt instructions (e.g., natural language steps or code templates), a contract couples visible skill instructions with an explicit \emph{contract runtime} that formalizes procedural bounds, including required phases, admissible actions, permitted transitions, and accepted termination.
By continuously inspecting agent actions against environment observations, the runtime dynamically tracks phase progression, blocks protocol violations, and guides valid next actions.

The contract runtime connects execution checking, training, and evaluation: it checks action admissibility, assigns credit for verified progress, and determines whether the prescribed procedure has been completed.
For training, we optimize the executor policy via reinforcement learning using GRPO~\bluecite{shao2024deepseekmath,li2025turnreward,lu2025environmenttuning} (Figure~\ref{fig:skill-perspectives}(c)).
To address sparse task rewards that appear only upon accepted termination, the runtime provides dense learning signals by assigning \emph{Verified Progress Credit} as contract milestones are reached, while offering \emph{Contract-State Feedback} during rollout to guide valid next steps.
For evaluation, the same runtime acts as an objective judge, recording \emph{Protocol Completion} when a trajectory satisfies all required phases and achieves accepted termination.
We report the resulting \textbf{Protocol Completion Rate (PCR)} by decoupling it from Task Outcome, thereby isolating procedural compliance from end-task performance.

We evaluate our approach using five executors across the Qwen2.5 and Qwen3.5 model families on Math and Search domains.
Contract training achieves remarkably high Protocol Completion Rates (PCR) across both domains.
Crucially, higher PCR does not uniformly translate into superior Task Outcomes, demonstrating that procedural compliance is largely decoupled from end-task performance.
Controlled studies and observation-withholding experiments analyze the individual contributions of visible skill instructions, training signals, observation content, and Contract-State Feedback.
Together, these findings establish that verifiable procedural execution with a fixed, expert-authored skill can be effectively trained and evaluated through a unified contract runtime.

Our main contributions are summarized as follows:
\begin{itemize}
\item \textbf{Formulation:} We formalize Grounded Skill-Following to achieve verifiable procedural execution on fixed expert skills across required phases, explicitly decoupling Protocol Completion from Task Outcome.
\item \textbf{Methodology:} We present a \emph{skill contract} runtime featuring \emph{Contract-State Feedback} for step-level action guidance and \emph{Verified Progress Credit} for dense reinforcement learning rewards.
\item \textbf{Findings:} Across two domains and five executors, we show that high Protocol Completion Rate can be trained without proportional gains in Task Outcome, revealing task-dependent impacts of instructions and observations.
\end{itemize}
\vspace{-3mm}
\section{Related Work}
\vspace{-2mm}
\subsection{From Instruction Following to Agent Skills}

Research on instruction compliance primarily targets response-level constraints within static contexts~\bluecite{wei2022finetuned,sanh2022multitask,wang2023selfinstruct,ouyang2022training}, measured by benchmarks that evaluate explicit and compositional requirements~\bluecite{zhou2023ifeval,wen2024complexbench}.
Tool-using architectures extend this setting to interactive environments, requiring models to interleave reasoning actions with observation feedback~\bluecite{schick2023toolformer,nakano2021webgpt,yao2023react}.
More recently, agent skills have emerged as a paradigm to package instructions, tool invocations, and execution strategies into structured, reusable procedures~\bluecite{anthropic2025skills,wang2023voyager,wang2024awm}.
Literature in this space heavily emphasizes skill discovery and retrieval~\bluecite{li2025sage,su2026skillretrieval}, skill evolution and composition~\bluecite{yang2026skillrl,zhang2026skill1,zhang2026skillcomposer}, or skill extraction and internalization into model weights~\bluecite{liu2026skiller,chen2026skill0,wang2026skillsd,lin2026skillc}.

Closest to our setting are SkillCoach and SKT~\bluecite{skillcoach2026,tan2026skt}, which evaluate or adapt provided skills to maximize task outcomes.
However, they allow skills to be modified or treat them as soft prompts, leaving phase-level execution unverified.
In contrast, \emph{grounded skill-following} treats an expert skill as a fixed, non-negotiable contract.
Instead of adapting the skill or taking outcome-driven shortcuts, our contract runtime enforces and evaluates strict phase-by-phase protocol compliance during agent-environment interaction.

\vspace{-2mm}
\subsection{Verifiable Control and Intermediate Supervision}

Verifiable control uses finite-state reward machines to structure long-horizon reinforcement learning~\bluecite{icarte2018rewardmachines}, while process supervision and turn-level rewards provide dense credit assignment for reasoning and tool-use trajectories~\bluecite{lightman2023verify,li2025turnreward,jin2025searchr1,deepseekai2025retool,lu2025environmenttuning}.
In parallel, constrained decoding forces output grammars during inference~\bluecite{ugare2024syncode}, dynamic interventions like HASP convert skills into execution rules~\bluecite{liu2026hasp}, and auditing methods motivate testing procedural execution independently from outcomes~\bluecite{ma2025proofofuse,huang2026harnessif,zhou2026counterfactual}.

Our approach differs fundamentally in how procedural specifications are formalized, trained, and evaluated.
Unlike reward machines that require rigid state definitions or constrained decoding that acts as an inference-only filter, our \emph{skill contract} pairs visible instructions with an explicit contract runtime to handle dynamic phase transitions.
Furthermore, while existing agentic RL remains primarily outcome-driven, our \emph{contract runtime} provides \emph{contract-state feedback} and \emph{verified progress credit} to optimize phase-by-phase compliance during training.
For evaluation, it decouples \emph{Protocol Completion Rate (PCR)} from \emph{Task Outcome} to measure procedural compliance independently.
\vspace{-2mm}
\section{Learning Grounded Skill-Following with Skill Contracts}
\vspace{-2mm}

\begin{figure*}[t]
\centering
\includegraphics[width=\textwidth]{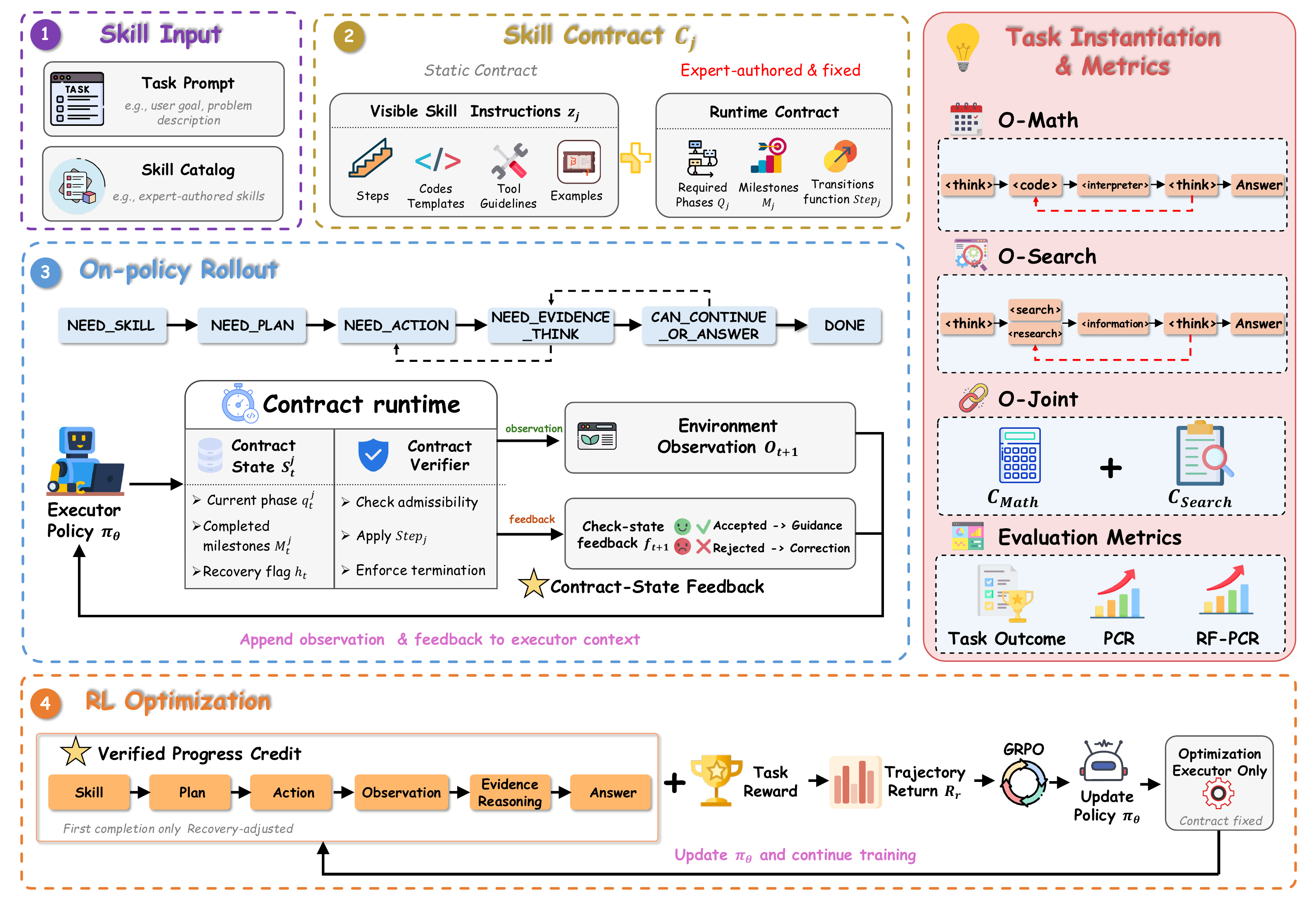}
\vspace{-8mm}
\caption{Skill-contract training for grounded skill-following. The contract $\mathcal C_j$ pairs visible skill instructions $z_j$ with a contract runtime. The runtime checks each action, credits newly verified progress, and provides contract-state feedback after each nonterminal transition. The feedback describes the current phase without exposing internal bookkeeping.}
\label{fig:contract}
\end{figure*}

Given a task and an expert-authored skill, we train an executor to follow the required procedure while grounding actions in environment observations.
A skill contract pairs visible skill instructions with an explicit contract runtime, providing Contract-State Feedback during rollout and Verified Progress Credit added to the trajectory return.
Figure~\ref{fig:contract} summarizes the framework.

\vspace{-2mm}
\subsection{Problem and Learning Requirement}

\paragraph{Contract representation and runtime.}
For skill $j$, we formalize the static skill contract as
\begin{equation}
\label{eq:contract-definition}
\mathcal C_j=(z_j,\mathcal Q_j,\mathcal M_j,\operatorname{Step}_j).
\end{equation}
Here, $z_j$ contains the visible skill instructions, $\mathcal Q_j$ is the set of required phases, $\mathcal M_j$ is the set of credit-bearing milestones, and $\operatorname{Step}_j$ is the transition function governing the contract runtime.
Operationalized during agent-environment interaction, the contract runtime integrates two core mechanisms:
\begin{itemize}
    \item \textbf{Dynamic Contract State ($S_t^j$):} Maintained at step $t$ as $S_t^j = (q_t^j, \mathcal M_t^j, h_t)$, comprising the current required phase $q_t^j \in \mathcal Q_j$, completed milestones $\mathcal M_t^j \subseteq \mathcal M_j$, and recovery flag $h_t \in \{0,1\}$. The current phase dictates admissible next actions and valid termination, while milestones track progress along the sequence.
    \item \textbf{Contract Verifier:} Implements the state transition function $\operatorname{Step}_j$ alongside a task adapter. The verifier checks action admissibility against environment observations, verifies phase transitions $S_t^j \to S_{t+1}^j$, and enforces accepted termination conditions.
\end{itemize}
Both visible skill instructions $z_j$ and the contract runtime logic are specified from the expert-authored skill (Appendix Tables~\ref{tab:contract-phases}--\ref{tab:runtime-adapters}).

\paragraph{Why Task Outcome alone is uninformative.}
GRPO samples $G$ trajectories for a prompt and normalizes their trajectory returns within the group~\bluecite{shao2024deepseekmath}.\ %
For trajectory return $R_i$, group standard deviation $s_R$, and a positive stabilizer $\epsilon$, the group-normalized advantage is
\begin{equation}
\label{eq:grpo-advantage}
\widehat A_i
=
\frac{R_i-\bar R}{s_R+\epsilon},
\qquad
\bar R=\frac{1}{G}\sum_{g=1}^{G}R_g.
\end{equation}
When trajectory returns depend solely on the binary Task Outcome, all trajectories in a group frequently share the same return, yielding $s_R=0$ and $\widehat A_i=0$ for all $i$.\ %
Under these zero-advantage conditions, the policy update provides no learning signal to distinguish among trajectories.\ %
Adding Verified Progress Credit to the trajectory return can break these zero-variance ties when trajectories accumulate different recovery-adjusted progress totals, even when Task Outcome is sparse or uniformly zero.\ %

\subsection{Contract-State Feedback and Verified Progress Credit}
\label{sec:contract-learning}

\paragraph{One contract transition.}
Let $x_t^j$ be the model-visible context before action $t$, and let $S_t^j = (q_t^j, \mathcal M_t^j, h_t)$ be the current contract state.
Interaction starts at $t=0$ with the initial state $S_0^j = (\texttt{NEED\_SKILL}, \varnothing, 0)$.
The initial context $x_0^j$ contains the system scaffold and task prompt, while visible skill instructions $z_j$ are appended upon the first accepted skill call.
At step $t$, the policy samples an action:
\begin{equation}
\label{eq:contract-action}
a_t \sim \pi_\theta(\cdot \mid x_t^j).
\end{equation}
The contract runtime then executes a single state transition via $\operatorname{Step}_j$:
\begin{equation}
\label{eq:contract-step}
(S_{t+1}^j, o_{t+1}, f_{t+1}) = \operatorname{Step}_j(S_t^j, a_t),
\end{equation}
where $S_{t+1}^j = (q_{t+1}^j, \mathcal M_{t+1}^j, h_{t+1})$.
Specifically, the contract verifier evaluates action $a_t$ against the current phase $q_t^j$.
If $a_t$ is inadmissible, the verifier rejects the action, leaves $q_t^j$ and $\mathcal M_t^j$ unchanged, sets $h_{t+1}=1$, sets $o_{t+1}=\bot$, and generates Contract-State Feedback $f_{t+1}$.
If $a_t$ is admissible, the verifier accepts the action, applies its action-specific phase transition, updates completed milestones $\mathcal M_{t+1}^j$, and resets $h_{t+1}=0$.
Accepted environment actions additionally trigger the task adapter to return environment observation $o_{t+1}$, whereas accepted non-environment actions set $o_{t+1}=\bot$; an accepted action need not advance the phase, as in the optional Search evidence-organization transition.
The recovery flag $h_t$ tracks whether actions follow rejected feedback and is used to scale credit for newly completed milestones.
The feedback message $f_{t+1}$ explicitly specifies action admissibility, the current required phase, and an admissible next action.
When Contract-State Feedback is disabled, $h_t$ remains zero and $f_{t+1}$ is omitted.
The next context is updated via concatenation:
\begin{equation}
\label{eq:contract-prefix}
x_{t+1}^j = x_t^j \oplus a_t \oplus
\begin{cases}
z_j \oplus f_{t+1}, & \text{if the skill call is accepted for the first time},\\
o_{t+1} \oplus f_{t+1}, & \text{otherwise},
\end{cases}
\end{equation}
where $\bot$ elements are omitted.
Accepted termination ends the rollout without trailing feedback.
Appendix~\ref{app:contract-runtime} details the complete runtime execution order.

\paragraph{Trajectory return.}
For trajectory $i$, let $S_{i,t}^j = (q_{i,t}^j, \mathcal M_{i,t}^j, h_{i,t})$ be the contract state before action $t$.\ %
The newly verified milestones at step $t$ are
\begin{equation}
\label{eq:new-milestones}
\Delta\mathcal M_{i,t}^j
=
\mathcal M_{i,t+1}^j \setminus \mathcal M_{i,t}^j.
\end{equation}
We call the reward assigned to these first completions \emph{Verified Progress Credit}.\ %
For milestone weight $w_m > 0$, recovery coefficient $\rho \in (0,1]$ ($\rho = 0.90$ in our main experiments), and total steps $T_i$, the trajectory return is
\begin{equation}
\label{eq:trajectory-return}
R_i
=
r_{\mathrm{task},i}
+
\sum_{t=0}^{T_i-1} \rho^{h_{i,t}} \sum_{m \in \Delta\mathcal M_{i,t}^j} w_m.
\end{equation}
The binary Task Outcome $r_{\mathrm{task},i} \in \{0,1\}$ equals 1 only if the trajectory reaches accepted termination with a correct final answer, and 0 otherwise.\ %
Each milestone contributes to credit at most once per trajectory.\ %
The recovery flag $h_{i,t} \in \{0,1\}$ specifically discounts progress credit completed immediately after a rejected action ($h_{i,t}=1$).\ %
When Task Outcomes tie across a sampled group, Verified Progress Credit distinguishes trajectories whenever their recovery-adjusted credit totals differ.\ %
Different milestone sets do not by themselves guarantee distinct totals because milestone weights and recovery discounts can coincide.\ %
Appendix~\ref{app:learning-signal} derives conditions under which Verified Progress Credit alters normalized advantages and explains why return variance alone is insufficient for a useful parameter gradient.\ %
Contract-State Feedback updates the model-visible context without altering trajectory return values.\ %
Appendix Table~\ref{tab:contract-reward} reports all exact milestone weights.\ %

\subsection{Optimization and Task Instantiation}

\paragraph{Policy optimization.}
We optimize the policy using a clipped GRPO objective~\bluecite{shao2024deepseekmath}.
For trajectory $i$, let $\mathcal{L}_i$ denote the set of trainable token positions and $\widehat{A}_i$ the trajectory-level group-normalized advantage.
The objective is
\begin{equation}
\label{eq:grpo-objective}
\mathcal{J}(\theta)
=
\frac{1}{G} \sum_{i=1}^{G} \frac{1}{|\mathcal{L}_i|} \sum_{\ell \in \mathcal{L}_i} \left[ \min\left( r_{i,\ell}(\theta)\widehat{A}_i, \operatorname{clip}\left( r_{i,\ell}(\theta), 1-\epsilon_-, 1+\epsilon_+ \right)\widehat{A}_i \right) - \beta d^{\mathrm{KL}}_{i,\ell} \right],
\end{equation}
where
\[
r_{i,\ell}(\theta) = \frac{\pi_\theta(a_{i,\ell} \mid x_{i,\ell})}{\pi_{\theta_{\mathrm{old}}}(a_{i,\ell} \mid x_{i,\ell})},
\]
$\beta$ is the KL penalty coefficient, and $\epsilon_-$ and $\epsilon_+$ specify the clipping bounds.
Here, $a_{i,\ell}$ is a generated token and $x_{i,\ell}$ its preceding context in trajectory $i$.
The term $d^{\mathrm{KL}}_{i,\ell}$ regularizes the policy relative to a frozen reference policy, with its estimator and reference initialization specified in Appendix~\ref{app:experimental-details}.

Visible skill instructions, environment observations, and Contract-State Feedback remain in context with a loss mask of zero, while all model-generated tokens in a trajectory share the single trajectory-level advantage $\widehat A_i$.
The method therefore uses trajectory-level credit assignment rather than exact action-level or token-level credit assignment.
The skill contract shapes the trajectory return, but neither Verified Progress Credit nor feedback guarantees a nonzero expected parameter gradient, convergence, or improved Task Outcome.

\paragraph{Task instantiation.}
Math and Search tasks use identical contract fields and optimizer specifications, while Appendix Table~\ref{tab:runtime-adapters} gives their task-specific actions, observations, and completion checks.
The runtime maintains task-specific bookkeeping used by $\operatorname{Step}_j$ (e.g., tool budget tracking and duplicate action checks), detailed in Appendix~\ref{app:contract-runtime}.
For Math, $\operatorname{Step}_j$ maps a code action executing Python to an interpreter observation and requires reasoning over that result before taking another action or producing the final answer, following program-aided and tool-integrated reasoning paradigms~\bluecite{gao2023pal,chen2023programthoughts,gou2024tora}.
For Search, $\operatorname{Step}_j$ maps the initial \texttt{<search>} and subsequent \texttt{<research>} actions to retrieval observations, following multi-turn retrieval reasoning~\bluecite{jin2025searchr1}.
Search evidence processing may also include an optional \texttt{<summary>} phase that organizes existing evidence without calling the retriever or producing a new observation.
After the summary, the policy may retrieve additional evidence or provide a final answer, motivated by prior work on evidence refinement and context summarization~\bluecite{shi2025autorefine,wu2025resum}.
In both tasks, a trajectory achieves \emph{Protocol Completion} when it satisfies all required phases $\mathcal{Q}_j$ and reaches accepted termination.
Each skill contract is written manually from the corresponding expert-authored skill; automatic conversion of arbitrary skill text is outside the present scope.
\section{Experiments}

\subsection{Experimental Setup}
\label{sec:experimental-setup}

\paragraph{Tasks and evaluation.}
We train on MATH~\bluecite{hendrycksmath2021}, GSM8K~\bluecite{cobbe2021training}, Natural Questions~\bluecite{kwiatkowski2019nq}, and HotpotQA~\bluecite{yang2018hotpotqa}.\ %
Evaluation additionally includes TriviaQA~\bluecite{joshi2017triviaqa}, PopQA~\bluecite{mallen2023trust}, 2WikiMultiHopQA~\bluecite{ho2020constructing}, MuSiQue~\bluecite{trivedi2022musique}, and Bamboogle~\bluecite{press2023measuring}.\ %
Math uses pass@1 after numerical normalization, while Search uses normalized exact match.\ %
Domain averages are computed uniformly across datasets within each domain.\ %

We report Task Outcome, PCR, and RF-PCR for skill-contract conditions.\ %
Task Outcome measures answer correctness, using accepted answers for contract-trained policies and the baseline extractors described in Appendix~\ref{app:experimental-details}.\ %
For $N$ evaluated trajectories, let $T_i$ denote the number of model actions in trajectory $i$, $q_{i,T_i}^j$ its final phase, and $h_{i,t}$ its recovery flag before action $t$.\ %
The contract enters \texttt{DONE} only when a trajectory completes all required phases and reaches accepted termination, so Protocol Completion Rate (PCR) is
\begin{equation}
\label{eq:pcr}
\mathrm{PCR}
=
\frac{100}{N}\sum_{i=1}^{N}
\mathbf 1\!\left[q_{i,T_i}^{j}=\texttt{DONE}\right].
\end{equation}
Under the runtime with feedback enabled, Recovery-free PCR (RF-PCR) additionally requires that the recovery flag never become one:
\begin{equation}
\label{eq:rf-pcr}
\mathrm{RF\text{-}PCR}
=
\frac{100}{N}\sum_{i=1}^{N}
\mathbf 1\!\left[q_{i,T_i}^{j}=\texttt{DONE}\right]
\mathbf 1\!\left[\sum_{t=0}^{T_i-1}h_{i,t}=0\right].
\end{equation}
Thus, RF-PCR cannot exceed PCR and is distinct from disabling feedback at inference.\ %
Protocol Completion does not guarantee a correct answer, so these measures remain separate.\ %
Appendix~\ref{app:experimental-details} provides dataset sizes and complete evaluation rules.\ %

\paragraph{Models.}
Qwen2.5-3B~\bluecite{yang2024qwen25} is the primary base executor for method analysis and controlled studies.\ %
Qwen3.5-4B-Base tests transfer to a newer base initialization~\bluecite{qwen2026qwen35}.\ %
Qwen2.5-3B-Instruct and Qwen3.5-4B test the same skill-contract design after instruction or general post-training.\ %
Qwen2.5-7B provides a same-family scale extension.\ %

\paragraph{Comparisons and training.}
Direct and Prompt evaluate frozen executors without and with the contract runtime.\ %
Recipe denotes matched ReTool-style and Search-R1 adaptations~\bluecite{deepseekai2025retool,jin2025searchr1}, Ours-S denotes separate O-Math and O-Search policies, and Ours-J denotes joint training.\ %
All online conditions use GRPO, and controlled studies preserve the data, task adapter, contract verifier, and task reward while changing only the named component.\ %
Search comparisons share one retriever and corpus~\bluecite{jin2025empiricalsearch}, while Appendix~\ref{app:experimental-details} records baseline provenance and Appendix Table~\ref{tab:training_setup} gives the complete settings.\ %

\noindent\textbf{Research questions.}
The analysis first establishes grounded skill-following, then examines training behavior and inference behavior.\ %
\begin{enumerate}
    \item \textbf{RQ1: Grounded skill-following.} Does skill-contract training produce high Protocol Completion and useful Task Outcomes across domains and executors?
    \item \textbf{RQ2: Training behavior.} How do verified progress credit, contract-state feedback, and the recovery coefficient affect learning?
    \item \textbf{RQ3: Inference behavior.} How does execution depend on visible skill instructions, observation content, and contract-state feedback?
\end{enumerate}

\begin{table}[!t]
\caption{Task Outcome, PCR, and RF-PCR on Math and Search (\%).\ %
RF-PCR excludes samples that receive rejected-action feedback.\ %
Direct evaluates frozen checkpoints without the contract runtime, while Prompt preloads the skill and uses the runtime.\ %
Recipe pairs task-specific adaptations, Ours-S trains each domain separately, Ours-J trains them jointly, and \textit{n/a} denotes an absent protocol metric.\ %
Bold marks the highest value in each metric column within each model block, including ties.\ %
Averages are unweighted, while \textsuperscript{tr} and \textsuperscript{ho} mark training and held-out datasets.}
\label{tab:main_results}
\centering
\scriptsize
\setlength{\tabcolsep}{0.65pt}
\renewcommand{\arraystretch}{0.96}
\begin{tabular*}{\textwidth}{@{\extracolsep{\fill}}lrrrrrrrrrrrrrrr@{}}
\toprule
& \multicolumn{5}{c}{Math} & \multicolumn{10}{c}{Search} \\
\cmidrule(lr){2-6}\cmidrule(lr){7-16}
Setting & MATH\textsuperscript{tr} & GSM\textsuperscript{tr} & Avg. & PCR & RF-PCR & NQ\textsuperscript{tr} & TQA\textsuperscript{ho} & Pop\textsuperscript{ho} & HQA\textsuperscript{tr} & 2W\textsuperscript{ho} & MuS\textsuperscript{ho} & Bam\textsuperscript{ho} & Avg. & PCR & RF-PCR \\
\midrule
\multicolumn{16}{@{}l}{\textbf{Qwen2.5-3B}} \\
Direct & 17.00 & 26.08 & 21.54 & \textit{n/a} & \textit{n/a} & 2.27 & 6.31 & 3.11 & 2.84 & 5.16 & 0.12 & 1.60 & 3.06 & \textit{n/a} & \textit{n/a} \\
Prompt & 0.16 & 0.00 & 0.08 & 0.16 & 0.00 & 0.08 & 0.14 & 0.14 & 0.09 & 0.25 & 0.00 & 0.00 & 0.10 & 0.20 & 0.00 \\
Recipe & 45.16 & \textbf{81.20} & 63.18 & \textit{n/a} & \textit{n/a} & 45.26 & \textbf{61.73} & 44.00 & \textbf{32.14} & 29.21 & 6.37 & 10.40 & 32.73 & \textit{n/a} & \textit{n/a} \\
Ours-S & \textbf{48.62} & 80.59 & \textbf{64.61} & \textbf{97.29} & \textbf{94.54} & \textbf{46.34} & 60.53 & \textbf{45.67} & 30.34 & \textbf{29.87} & \textbf{6.91} & 16.00 & \textbf{33.67} & \textbf{99.82} & \textbf{98.54} \\
Ours-J & 45.02 & 79.22 & 62.12 & 94.46 & 93.78 & 43.23 & 58.23 & 43.96 & 29.09 & 28.53 & 5.13 & \textbf{18.40} & 32.37 & 98.46 & 97.00 \\
\midrule
\multicolumn{16}{@{}l}{\textbf{Qwen3.5-4B-Base}} \\
Direct & 70.60 & 88.93 & 79.77 & \textit{n/a} & \textit{n/a} & 12.96 & 35.18 & 12.87 & 15.65 & 22.65 & 2.69 & 20.80 & 17.54 & \textit{n/a} & \textit{n/a} \\
Prompt & 2.16 & 4.32 & 3.24 & 3.84 & 0.00 & 0.89 & 5.41 & 1.92 & 1.80 & 1.33 & 0.21 & 3.20 & 2.11 & 7.32 & 0.00 \\
Recipe & \textbf{78.28} & \textbf{95.22} & \textbf{86.75} & \textit{n/a} & \textit{n/a} & 48.59 & 65.34 & 47.06 & 38.46 & 35.97 & 12.78 & 31.20 & 39.91 & \textit{n/a} & \textit{n/a} \\
Ours-S & 72.02 & 91.81 & 81.92 & \textbf{98.92} & 96.12 & \textbf{49.39} & \textbf{65.75} & \textbf{48.28} & \textbf{47.13} & \textbf{46.91} & \textbf{20.19} & \textbf{44.80} & \textbf{46.06} & 99.76 & \textbf{99.11} \\
Ours-J & 71.86 & 93.40 & 82.63 & 98.73 & \textbf{96.31} & 48.42 & 63.18 & 45.92 & 41.36 & 36.80 & 14.52 & 38.40 & 41.23 & \textbf{99.99} & 95.37 \\
\midrule
\multicolumn{16}{@{}l}{\textbf{Qwen2.5-3B-Instruct}} \\
Direct & 48.26 & 77.79 & 63.02 & \textit{n/a} & \textit{n/a} & 8.86 & 21.52 & 8.71 & 13.49 & 22.49 & 1.32 & 3.20 & 11.37 & \textit{n/a} & \textit{n/a} \\
Prompt & 2.72 & 4.09 & 3.41 & 4.83 & 0.39 & 0.17 & 0.53 & 0.29 & 0.18 & 0.14 & 0.04 & 0.00 & 0.19 & 75.07 & 33.28 \\
Recipe & 47.16 & \textbf{83.40} & 65.28 & \textit{n/a} & \textit{n/a} & \textbf{45.10} & \textbf{61.43} & \textbf{44.24} & \textbf{32.06} & 27.72 & \textbf{7.41} & \textbf{14.40} & \textbf{33.19} & \textit{n/a} & \textit{n/a} \\
Ours-S & \textbf{51.24} & 80.29 & \textbf{65.76} & \textbf{98.83} & \textbf{96.54} & 44.99 & 60.62 & 43.94 & 31.74 & \textbf{28.58} & 7.03 & \textbf{14.40} & 33.04 & 99.92 & \textbf{99.69} \\
Ours-J & 45.34 & 80.29 & 62.81 & 96.72 & 91.12 & 42.99 & 59.78 & 43.22 & 29.95 & 27.77 & 6.37 & 9.60 & 31.38 & \textbf{99.97} & 99.41 \\
\midrule
\multicolumn{16}{@{}l}{\textbf{Qwen3.5-4B}} \\
Direct & 38.78 & 72.10 & 55.44 & \textit{n/a} & \textit{n/a} & 11.27 & 35.01 & 11.54 & 7.87 & 2.32 & 0.99 & 17.60 & 12.37 & \textit{n/a} & \textit{n/a} \\
Prompt & 0.38 & 1.29 & 0.83 & 0.93 & 0.00 & 0.47 & 6.51 & 1.55 & 1.90 & 1.23 & 0.29 & 2.40 & 2.05 & 5.71 & 0.00 \\
Recipe & 75.44 & \textbf{94.92} & 85.18 & \textit{n/a} & \textit{n/a} & 48.56 & 65.22 & \textbf{49.98} & 37.14 & 40.86 & 19.97 & 23.20 & 40.70 & \textit{n/a} & \textit{n/a} \\
Ours-S & \textbf{78.48} & 92.72 & \textbf{85.60} & 98.25 & \textbf{96.27} & 48.59 & 64.49 & 47.36 & \textbf{47.05} & \textbf{46.10} & 19.03 & 41.60 & 44.89 & 99.50 & \textbf{96.58} \\
Ours-J & 75.76 & 90.14 & 82.95 & \textbf{99.27} & 92.90 & \textbf{50.86} & \textbf{66.15} & 47.56 & 46.87 & 45.01 & \textbf{22.63} & \textbf{47.20} & \textbf{46.61} & \textbf{99.96} & 95.40 \\
\midrule
\multicolumn{16}{@{}l}{\textbf{Qwen2.5-7B}} \\
Direct & 15.70 & 23.96 & 19.83 & \textit{n/a} & \textit{n/a} & 5.48 & 12.72 & 4.35 & 5.51 & 9.05 & 1.41 & 5.60 & 6.30 & \textit{n/a} & \textit{n/a} \\
Prompt & 2.06 & 4.09 & 3.08 & 5.17 & 1.28 & 0.75 & 1.78 & 0.95 & 0.55 & 1.30 & 0.12 & 1.60 & 1.01 & 12.36 & 3.22 \\
Recipe & \textbf{58.96} & 86.42 & 72.69 & \textit{n/a} & \textit{n/a} & 45.83 & \textbf{66.38} & 45.82 & \textbf{37.14} & 32.29 & 8.13 & 24.00 & \textbf{37.08} & \textit{n/a} & \textit{n/a} \\
Ours-S & 58.72 & \textbf{86.73} & \textbf{72.73} & \textbf{96.08} & \textbf{95.50} & 43.63 & 64.28 & \textbf{46.93} & 35.56 & \textbf{33.19} & 10.05 & \textbf{25.60} & 37.03 & \textbf{98.28} & \textbf{98.27} \\
Ours-J & 56.96 & 83.40 & 70.18 & 96.00 & 95.40 & \textbf{47.81} & 59.01 & 42.43 & 34.21 & 30.44 & \textbf{16.79} & 24.00 & 36.38 & 97.00 & 96.67 \\
\bottomrule
\end{tabular*}
\end{table}

\subsection{RQ1: Grounded Skill-Following Results}
\label{sec:main-results}

\textbf{Contract training yields high Protocol Completion with mixed Task Outcome gains.}\ %
Across separate and joint policies, Math PCR ranges from 94.46\% to 99.27\%, while Search PCR ranges from 97.00\% to 99.99\% (Table~\ref{tab:main_results}).\ %
For Qwen3.5-4B, joint training achieves Math/Search Task Outcomes of 82.95\%/46.61\%, PCR of 99.27\%/99.96\%, and RF-PCR of 92.90\%/95.40\%.\ %
The gap between PCR and RF-PCR is configuration-dependent: for example, Qwen3.5-4B joint training reaches 99.27\% Math PCR but 92.90\% RF-PCR, whereas Qwen2.5-7B separate Math training reaches 96.08\% PCR and 95.50\% RF-PCR.\ %
Task Outcome does not uniformly improve over Recipe, nor does joint training consistently outperform separate training.\ %
Appendix~\ref{app:execution-traces} provides complete execution traces.\ %

\subsection{RQ2: Training Behavior}
\label{sec:training-behavior}

\paragraph{Training signals.}
\textbf{Verified progress credit drives learning, with task-dependent support from feedback.}\ %
Credit-only reaches Math/Search Task Outcomes of 59.76\%/29.28\% and PCR of 93.17\%/98.69\%, while Feedback + Credit reaches 64.61\%/30.11\% and 97.29\%/98.90\%, respectively (Figure~\ref{fig:contract-acquisition}(a), Appendix Table~\ref{tab:codesign_ablation}).\ %
Outcome-only and Feedback-only remain at zero in these conditions.\ %
For the matched learning curves, Outcome-only and Credit-only use the same contract runtime with contract-state feedback disabled.\ %
Outcome-only uses the task reward alone, while Credit-only adds verified progress credit to the trajectory return.\ %
Credit-only develops nonzero Task Outcome and PCR while Outcome-only remains at zero under this contract runtime (Figure~\ref{fig:acquisition-dynamics}).\ %

\begingroup
\setlength{\intextsep}{8pt}
\begin{figure}[H]
\centering
\includegraphics[width=0.90\textwidth]{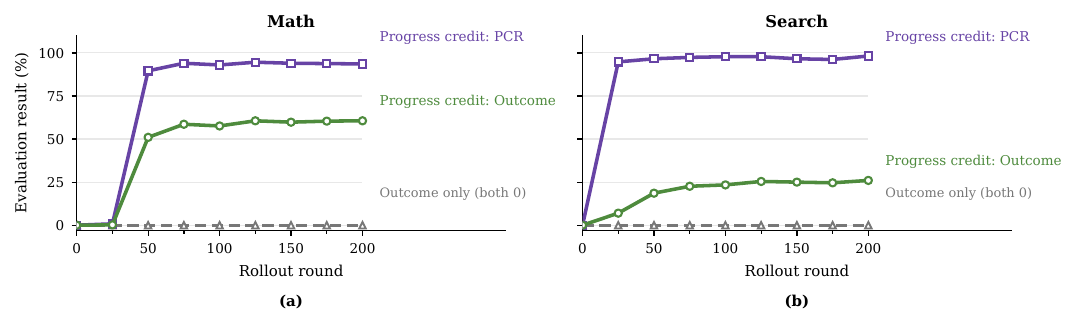}
\vspace{-4mm}
\caption{Qwen2.5-3B learning dynamics under the same contract runtime with feedback disabled.\ %
Each point averages a fixed monitor of 250 examples per dataset: (a) MATH and GSM8K, (b) Natural Questions and MuSiQue.\ %
Verified progress credit yields nonzero PCR and Task Outcome in both domains, while Outcome-only remains at zero.}
\label{fig:acquisition-dynamics}
\end{figure}
\endgroup

\paragraph{Recovery coefficient.}
\textbf{The recovery coefficient of 0.90 yields higher RF-PCR in the tested joint conditions.}\ %
With $\rho=0.90$, Math/Search RF-PCR is 93.78\%/98.18\%, compared with 0.00\%/88.33\% for $\rho=1.00$ (Table~\ref{tab:runtime-recovery-boundary}, Panel A).\ %
The $\rho=1.00$ condition has higher PCR and Task Outcome, with Math/Search Task Outcome higher by 2.49/0.70 points.\ %
These independently trained conditions show differences in recovery-free Protocol Completion, without establishing a consistent advantage across metrics or training seeds.\ %

\begin{table}[!htbp]
\vspace{-4mm}
\caption{Training- and inference-time recovery interventions.\ %
Panel A compares independently trained Qwen2.5-3B joint conditions with $\rho=0.90$ and $\rho=1.00$.\ %
Panel B removes feedback at inference from frozen Qwen2.5-3B single-task checkpoints.\ %
Both panels report unweighted averages over MATH and GSM8K for Math and 2WikiMultiHopQA and HotpotQA for Search.\ %
Values are percentages, and changes are percentage points.}
\label{tab:runtime-recovery-boundary}
\centering
\footnotesize
\setlength{\tabcolsep}{2.5pt}
\renewcommand{\arraystretch}{1.15}
\begin{tabular*}{\textwidth}{@{\extracolsep{\fill}}lrrrr@{\hspace{12pt}}rrrrr@{}}
\toprule
& \multicolumn{4}{c}{\textbf{A: Recovery coefficient}} & \multicolumn{5}{c}{\textbf{B: Feedback removal}} \\
\cmidrule(lr){2-5}\cmidrule(lr){6-10}
\multirow{2}{*}{Domain} & \multirow{2}{*}{$\rho$} & \multirow{2}{*}{\shortstack{Task\\Outcome}} & \multirow{2}{*}{PCR} & \multirow{2}{*}{RF-PCR} & \multicolumn{3}{c}{Task Outcome} & \multicolumn{2}{c}{PCR} \\
\cmidrule(lr){6-8}\cmidrule(lr){9-10}
& & & & & On & Off & Change & On & Off \\
\midrule
\multirow{2}{*}{Math} & 0.90 & 62.12 & 94.46 & 93.78 & \multirow{2}{*}{64.61} & \multirow{2}{*}{58.25} & \multirow{2}{*}{$-6.36$} & \multirow{2}{*}{97.29} & \multirow{2}{*}{86.93} \\
& 1.00 & 64.61 & 96.23 & 0.00 & & & & & \\
\midrule
\multirow{2}{*}{Search} & 0.90 & 28.81 & 98.90 & 98.18 & \multirow{2}{*}{30.11} & \multirow{2}{*}{13.15} & \multirow{2}{*}{$-16.96$} & \multirow{2}{*}{98.90} & \multirow{2}{*}{43.05} \\
& 1.00 & 29.51 & 99.53 & 88.33 & & & & & \\
\bottomrule
\end{tabular*}
\vspace{-2mm}
\end{table}

\subsection{RQ3: Inference Behavior}
\label{sec:inference-behavior}

We intervene on frozen Qwen2.5-3B single-task policies evaluated on MATH, GSM8K, 2WikiMultiHopQA, and HotpotQA.\ %
Full intervention conditions are reported in Appendix~\ref{app:controlled-results}.\ %

\begin{figure}[H]
\centering
\includegraphics[width=\textwidth]{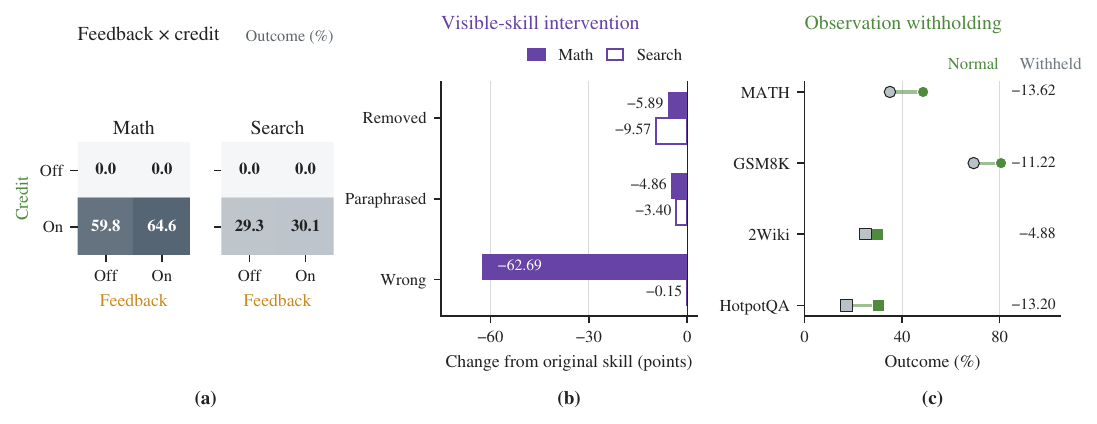}
\vspace{-8mm}
\caption{Qwen2.5-3B interventions on (a) contract-state feedback and verified progress credit during training, (b) visible skill instructions at inference, and (c) observation content at inference.\ %
Values are percentages or percentage-point changes.}
\label{fig:contract-acquisition}
\end{figure}

\paragraph{Visible skill instructions.}
\textbf{Execution depends on visible skill instructions.}\ %
Removing them lowers Math/Search Task Outcome by 5.89/9.57 points, while paraphrasing preserves most performance (Figure~\ref{fig:contract-acquisition}(b), Appendix Table~\ref{tab:skill_counterfactual}).\ %
With the rest of $\mathcal C_j$ fixed, this tests instruction dependence within the runtime.\ %
Wrong-skill substitution also changes the action grammar, limiting its interpretation to compatibility.\ %

\paragraph{Observation and feedback dependence.}
\textbf{Observation content supports Task Outcome, while feedback dependence is stronger on Search.}\ %
Withholding observation content lowers Task Outcome by 13.62, 11.22, 4.88, and 13.20 points on MATH, GSM8K, 2Wiki, and HotpotQA, respectively (Figure~\ref{fig:contract-acquisition}(c)).\ %
Removing feedback lowers Math/Search Task Outcome by 6.36/16.96 points (Table~\ref{tab:runtime-recovery-boundary}, Panel B).\ %
PCR falls from 97.29\% to 86.93\% on Math and from 98.90\% to 43.05\% on Search.\ %
These inference interventions establish behavioral dependence, without isolating token-level evidence use or feedback's effect during training.\ %

\vspace{-2mm}

\vspace{-4mm}
\subsection{Empirical Insights}

\textbf{High Protocol Completion does not imply improved Task Outcome.}\ %
Trained executors achieve high PCR, but their Task Outcome gains over matched Recipe vary (Section~\ref{sec:main-results}).\ %
\textbf{High Protocol Completion can coexist with different execution dependencies.}\ %
Math is more sensitive to cross-task skill substitution, whereas Search is more sensitive to feedback removal, although substitution also changes the action grammar (Section~\ref{sec:inference-behavior}).\ %
\textbf{Small training-time gains do not imply feedback independence.}\ %
On Search, adding feedback during training yields a small Task Outcome increment, yet removing it at inference substantially lowers Task Outcome and PCR (Sections~\ref{sec:training-behavior} and~\ref{sec:inference-behavior}).\ %
Feedback can guide subsequent attempts or output refinement, as also explored in Reflexion and CRITIC~\bluecite{shinn2023reflexion,gou2023critic}.\ %

\vspace{-2mm}
\section{Conclusion}
\vspace{-2mm}

We define grounded skill-following as learning to execute a fixed expert-authored skill through its required phases and to use environment observations to complete the task.\ %
Skill contracts make procedural progress verifiable during post-training and evaluation.\ %
Math and Search experiments report PCR, RF-PCR, and Task Outcome separately, while controlled learning curves show nonzero Protocol Completion and Task Outcome with verified progress credit under a contract runtime where outcome-only training remains at zero.\ %
Reliance on visible skill instructions and contract-state feedback varies by task, and the findings remain limited to manually specified contracts and the tested executors (Appendix~\ref{app:discussion}).\ %

\section*{AI Use Statement}
We used generative AI tools to assist with writing and polishing, including drafting and revising sections of this manuscript.\ %
AI tools also supported literature retrieval and discovery, including identifying relevant prior work.\ %
We used these tools for research ideation and experimental execution, as well as assistance in developing proofs of mathematical claims.\ %
The authors take full responsibility for the final content, including the references, mathematical claims and proofs, experimental results, analyses, and text.\ %

\section*{Reproducibility Statement}

Appendix A specifies the contract runtime, transition logic, milestone definitions, and reward construction. Appendix B provides the exact model-visible skill interfaces. Appendix C reports training configurations, optimization hyperparameters, rollout budgets, retrieval settings, dataset sizes, and evaluation procedures. Appendix D provides complete representative execution traces, and Appendix E specifies the controlled intervention settings and results. The code and implementation artifacts will be made publicly available after consolidation and documentation.

\bibliographystyle{iclr2027_conference}
\bibliography{references}

\clearpage
\appendix
\section{Skill Contract Runtime and Learning Signal}

\subsection{Canonical Contract Runtime}
\label{app:contract-runtime}

The contract runtime maintains the contract state $S_t^j = (q_t^j, \mathcal M_t^j, h_t)$, updating the current phase $q_t^j$ and completed milestones $\mathcal M_t^j$ alongside environment observations $o_{t+1}$ and contract-state feedback $f_{t+1}$ after each action.
The symbol $\bot$ denotes the absence of an environment observation.
The executor receives feedback regarding the contract phase rather than internal bookkeeping details.
The recovery flag $h_t$ is initialized to zero and is set to $1$ upon receiving feedback from a rejected action.
When an action is accepted, any newly verified milestones are recorded using the pre-action recovery flag $h_t$, after which $h_{t+1}$ is reset to $0$ (even if no new milestone is completed).
When contract-state feedback is disabled, the recovery flag remains zero throughout the rollout.
The runtime also tracks task-specific bookkeeping omitted from Equation~\ref{eq:contract-step}.
Observation validity is initially false and is updated after each environment call to evaluate phase transitions following evidence reasoning.
Environment-action counts enforce tool-call budgets and distinguish initial from subsequent retrieval actions; the most recent query prevents duplicate retries, and a record of evidence organization prevents redundant summary phases.

\begin{table}[H]
\caption{Phase set $\mathcal Q_j$ shared by Math and Search. A rejected action preserves the current phase and returns one admissible next action. Table~\ref{tab:runtime-adapters} defines substantive thought.}
\label{tab:contract-phases}
\centering
\footnotesize
\setlength{\tabcolsep}{3pt}
\renewcommand{\arraystretch}{1.08}
\begin{tabularx}{\linewidth}{@{}>{\raggedright\arraybackslash}p{0.25\linewidth}>{\raggedright\arraybackslash}p{0.27\linewidth}>{\raggedright\arraybackslash}X@{}}
\toprule
Phase & Accepted action & Next phase \\
\midrule
\texttt{NEED\_SKILL} & Expected skill call & \texttt{NEED\_PLAN} \\
\texttt{NEED\_PLAN} & Substantive planning thought & \texttt{NEED\_ACTION} \\
\texttt{NEED\_ACTION} & Admissible environment action & \texttt{NEED\_EVIDENCE\_\allowbreak{}THINK} \\
\texttt{NEED\_EVIDENCE\_\allowbreak{}THINK} & Substantive evidence reasoning & \texttt{CAN\_CONTINUE\_\allowbreak{}OR\_ANSWER} after a valid observation, otherwise \texttt{NEED\_ACTION} \\
\texttt{CAN\_CONTINUE\_\allowbreak{}OR\_ANSWER} & Repeated environment action, first optional Search evidence-organization action with \texttt{<summary>}, or final answer & \texttt{NEED\_EVIDENCE\_\allowbreak{}THINK}, \texttt{CAN\_CONTINUE\_\allowbreak{}OR\_ANSWER}, or \texttt{DONE}, respectively. A repeated evidence-organization action is rejected \\
\texttt{DONE} & None & Terminal \\
\bottomrule
\end{tabularx}
\end{table}

\FloatBarrier

\begin{table}[H]
\caption{Runtime execution order for one skill-contract trajectory.}
\label{tab:canonical_procedure}
\centering
\small
\setlength{\tabcolsep}{4pt}
\renewcommand{\arraystretch}{1.08}
\begin{tabularx}{\linewidth}{@{}cl>{\raggedright\arraybackslash}X@{}}
\toprule
Order & Operation & Update \\
\midrule
1 & Initialize & Set initial contract state $S_0^j = (\texttt{NEED\_SKILL}, \varnothing, 0)$. Initialize context $x_0^j$ with system scaffold and task prompt (Appendix~\ref{app:exact-prompts-skills}). \\
2 & Generate & Sample action $a_t \sim \pi_\theta(\cdot \mid x_t^j)$. \\
3 & Transition & Execute state transition via Equation~\ref{eq:contract-step}. Set $o_{t+1}=\bot$ when no observation is generated. Action rejection preserves $q_t^j$ and $\mathcal M_t^j$, setting $h_{t+1}=1$ when feedback is enabled. Action acceptance records new milestones scaled by $\rho^{h_t}$, then resets $h_{t+1}=0$. \\
4 & Record & Record newly verified milestones $\Delta\mathcal M_{i,t}^j$ and update context via Equation~\ref{eq:contract-prefix}, omitting $\bot$ and prepending skill instructions $z_j$ upon first accepted skill call. \\
5 & Repeat & Repeat Order 2--4 until accepted termination or budget exhaustion. Exhaustion does not constitute Protocol Completion. \\
6 & Score & Compute final trajectory return $R_i$ using Equation~\ref{eq:trajectory-return}. \\
\bottomrule
\end{tabularx}
\end{table}
\FloatBarrier

Each model turn is truncated after its first complete leading action.
The environment responds in a separate loss-masked turn.
A rejected action remains trainable, whereas visible skill instructions, contract-state feedback, and environment observations are assigned a loss mask of zero.
An inadmissible action does not advance the phase or milestone set, terminate the rollout, or impose a negative penalty.
The executor may continue generating actions until accepted termination or budget depletion.
Training on full interaction trajectories is motivated by research on learning to recover from unsuccessful attempts~\bluecite{qu2024rise}.
Our evaluation assesses this trajectory-level formulation without isolating the individual gradient contributions of rejected-action tokens.

\paragraph{Validity of the accepted procedure.}
Assume that the initial contract state $S_0^j = (q_0^j, \mathcal M_0^j, h_0)$ satisfies the contract specification and that the contract verifier accepts an action $a_t$ only when its induced transition $S_t^j \to S_{t+1}^j$ is valid.
Because action rejection leaves the current phase $q_t^j$ and completed milestones $\mathcal M_t^j$ unchanged, it trivially preserves state validity.
Each accepted action preserves validity by definition, including optional Search evidence-organization transitions that maintain the current phase.
By mathematical induction over contract transitions, any accepted sequence of transitions strictly satisfies the procedure specified by the contract along its traversed phases.
When accepted termination requires satisfying all phases in $\mathcal Q_j$, Protocol Completion Rate (PCR) precisely measures this structural property.

This property guarantees structural adherence under valid verifier logic; it does not guarantee that the executor will reach termination.
It neither validates raw rejected actions nor guarantees answer correctness or meaningful semantic processing of observations.
Furthermore, removing rejected actions and replaying the remaining action sequence does not reproduce the original trajectory, as rejections directly alter contract-state feedback $f_{t+1}$ and recovery flag bookkeeping $h_t$.

\subsection{Feedback, Milestones, and Task Adapters}
\label{app:feedback-reward}

Each contract-state feedback message $f_{t+1}$ explicitly specifies action admissibility, the current phase, and an admissible next action.
It never exposes the ground-truth task answer.
The contract runtime prepends a status prefix to a phase-specific message body, wrapping the combined message in \texttt{<hint>...</hint>}.

\begin{lstlisting}[style=feedbackcode]
NEED_SKILL:
State: no skill is loaded. Next: select one listed skill with <skill_call>...</skill_call>.

NEED_PLAN:
State: the skill is loaded. Next: write one complete <think>...</think> plan.

NEED_ACTION, Math:
State: Python evidence is required. Next: run calculation in <code>...</code> and print a result.

NEED_ACTION, first Search action:
State: retrieval evidence is required. Next: use a concise <search>...</search> query.

NEED_ACTION, later Search action:
State: retrieval evidence is required. Next: use a concise <research>...</research> query.

NEED_EVIDENCE_THINK:
State: an environment observation is ready. Next: write one complete <think>...</think>.

CAN_CONTINUE_OR_ANSWER, Math:
State: evidence is analyzed. Next: use <code>...</code> again or give Answer: \boxed{...}.

CAN_CONTINUE_OR_ANSWER, Search before summary:
State: evidence is analyzed. Next: use <research>...</research> again, organize it with <summary>...</summary>, or give <answer>...</answer>.

CAN_CONTINUE_OR_ANSWER, Search after summary:
State: evidence is summarized. Next: use <research>...</research> again or give <answer>...</answer>.
\end{lstlisting}

The status prefixes are \texttt{Action accepted.} and \texttt{Action rejected.}.
An accepted final answer terminates the rollout immediately without appended feedback.
In ablation studies where contract-state feedback is disabled, $f_{t+1}$ is omitted after both accepted and rejected actions while contract validation logic remains fully active.

For Math, a valid observation $o_{t+1}$ requires successful Python code execution with non-empty stdout.
For Search, a valid observation requires at least one non-empty retrieved document.
An admissible environment action transitions to evidence reasoning (\texttt{NEED\_EVIDENCE\_THINK}) even if the returned observation is invalid.
Following evidence reasoning, the task adapter permits state transition to \texttt{CAN\_CONTINUE\_OR\_ANSWER} only if the preceding observation was valid; otherwise, it requires an additional environment action.
These checks verify structural non-emptiness rather than semantic relevance or sufficiency.

Search additionally permits one optional evidence-organization action following evidence reasoning.
This action uses \texttt{<summary>}, preserves the \texttt{CAN\_CONTINUE\_OR\_ANSWER} phase, generates no environment observation ($o_{t+1}=\bot$), and earns verified progress credit only upon its first acceptance.
Its content check enforces a minimal length requirement of at least 24 normalized characters and four word tokens without embedded XML tags, preventing empty or trivial summaries.

\begin{table}[H]
\caption{Milestone set $\mathcal M_j$ and first-completion weights used for verified progress credit. Each contract location maps the credit event to the shared phase or transition in Table~\ref{tab:contract-phases}. The task reward is a separate binary term. Table~\ref{tab:runtime-adapters} defines substantive thought.\ %
For Search, evidence organization through \texttt{<summary>} is optional for accepted termination.\ %
Its verified progress credit encourages this optional action, so permitted trajectories need not receive equal progress credit.}
\label{tab:contract-reward}
\centering
\footnotesize
\setlength{\tabcolsep}{2pt}
\renewcommand{\arraystretch}{1.08}
\begin{tabularx}{\linewidth}{@{}>{\raggedright\arraybackslash}p{0.14\linewidth} l c >{\raggedright\arraybackslash}p{0.27\linewidth} >{\raggedright\arraybackslash}X@{}}
\toprule
Milestone & Scope & Weight & Contract location & First-completion condition \\
\midrule
Skill & Both & 0.10 & \texttt{NEED\_SKILL} & The expected \texttt{<skill\_call>} is accepted. \\
Plan & Both & 0.10 & \texttt{NEED\_PLAN} & A substantive \texttt{<think>} is accepted. \\
Action & Both & 0.20 & \texttt{NEED\_ACTION} & The first admissible \texttt{<code>}, \texttt{<search>}, or \texttt{<research>} passes its task adapter. \\
Observation & Both & 0.20 & \texttt{NEED\_ACTION} $\rightarrow$ \texttt{NEED\_\allowbreak{}EVIDENCE\_\allowbreak{}THINK} & Math execution produces nonempty output, or Search returns at least one nonempty document. \\
Evidence reasoning & O-Math & 0.20 & \texttt{NEED\_\allowbreak{}EVIDENCE\_\allowbreak{}THINK} & A substantive \texttt{<think>} is accepted. \\
Evidence reasoning & O-Search & 0.10 & \texttt{NEED\_\allowbreak{}EVIDENCE\_\allowbreak{}THINK} & A substantive \texttt{<think>} is accepted. \\
Evidence organization & O-Search & 0.10 & \texttt{CAN\_\allowbreak{}CONTINUE\_\allowbreak{}OR\_\allowbreak{}ANSWER} $\rightarrow$ same phase & The first optional \texttt{<summary>} passes the content checks after evidence reasoning without calling the environment. \\
Answer & Both & 0.20 & \texttt{CAN\_\allowbreak{}CONTINUE\_\allowbreak{}OR\_\allowbreak{}ANSWER} $\rightarrow$ \texttt{DONE} & A nonempty boxed Math answer or tagged Search answer is accepted. \\
\bottomrule
\end{tabularx}

\vspace{2pt}
\parbox{0.97\linewidth}{At trajectory end, each milestone contributes at most once to $R_i$. A milestone first completed by the next accepted transition after rejected-action feedback contributes $\rho w_m$. Otherwise, it contributes $w_m$.\par\smallskip The main runs use $\rho=0.90$. The registered no-discount comparison independently trains a condition with $\rho=1.00$. The coefficient does not modify earlier progress or the independent task reward.\par\smallskip O-Search retains separate 0.10 first-completion credits for evidence reasoning and optional evidence organization through \texttt{<summary>}.}
\end{table}

\FloatBarrier

\subsection{Motivation for Group Return Variance under Sparse Rewards}
\label{app:learning-signal}

For a given task prompt, let $R_1, \ldots, R_G$ be the trajectory returns of $G$ sampled rollouts, with empirical mean $\bar R$ and standard deviation $s_R$.
Let $r_{\mathrm{task},i}$ denote the sparse task-completion reward and $p_i$ denote the total recovery-adjusted progress credit defined in Equation~\ref{eq:trajectory-return}.
The trajectory return decomposes as:
\begin{equation}
\label{eq:return-decomposition}
R_i = r_{\mathrm{task},i} + p_i.
\end{equation}

When all rollouts within a sampled group fail to solve the task (or all succeed), the task reward $r_{\mathrm{task},i}$ is uniform across the group.
In this all-zero (or all-one) reward regime, the pairwise return difference reduces solely to differences in verified progress credit:
\begin{equation}
\label{eq:return-difference}
R_i - R_k = p_i - p_k.
\end{equation}
If at least two trajectories have different recovery-adjusted progress-credit totals $p_i \neq p_k$, their returns differ and yield a non-zero empirical variance $s_R > 0$.
This can occur when trajectories complete different milestones or incur different recovery discounts, but different milestone sets alone do not guarantee distinct totals.
This allows the normalized advantages in Equation~\ref{eq:grpo-advantage} to provide meaningful learning signals for step-by-step progress, even before correct final answers become frequent.
Figure~\ref{fig:acquisition-dynamics} illustrates training dynamics under these reward formulations.

Conversely, if every trajectory receives an identical progress credit $c$, group mean centering cancels this additive offset, where $\bar r_{\mathrm{task}}$ denotes the group mean task reward:
\begin{equation}
\label{eq:common-credit-cancels}
R_i = r_{\mathrm{task},i} + c
\quad\Longrightarrow\quad
R_i - \bar R = r_{\mathrm{task},i} - \bar r_{\mathrm{task}}.
\end{equation}
Because standard deviation remains invariant under constant shifts ($s_{R+c} = s_R$), a uniform progress credit leaves normalized GRPO advantages completely unchanged.

\begin{table}[H]
\caption{Task-specific instantiation of $\operatorname{Step}_j$ by phase. Table~\ref{tab:contract-phases} defines the shared transitions, and Table~\ref{tab:contract-reward} locates verified progress credit.\ %
For \texttt{<think>}, ``substantive'' requires at least 12 characters after whitespace normalization, including at least 6 alphanumeric characters.\ %
These are form-based checks and do not assess reasoning quality.}
\label{tab:runtime-adapters}
\centering
\footnotesize
\setlength{\tabcolsep}{2.5pt}
\renewcommand{\arraystretch}{1.07}
\begin{tabularx}{\linewidth}{@{}>{\raggedright\arraybackslash}p{0.25\linewidth}>{\raggedright\arraybackslash}X>{\raggedright\arraybackslash}X@{}}
\toprule
Phase & O-Math & O-Search \\
\midrule
\texttt{NEED\_SKILL} & Accept the expected \texttt{code-interpreter-protocol} skill call. & Accept the expected \texttt{qa-search-protocol} skill call. \\
\texttt{NEED\_PLAN} & Accept a substantive \texttt{<think>} that plans the calculation. & Accept a substantive \texttt{<think>} that plans retrieval. \\
\texttt{NEED\_ACTION} & Accept executable Python that prints a result in \code{}.
The restricted interpreter returns \texttt{<interpreter>}.
The observation is valid when execution succeeds and printed output is nonempty.
The accepted action enters \texttt{NEED\_EVIDENCE\_THINK}. & Accept the first admissible query in \texttt{<search>} and later nonrepeated admissible queries in \texttt{<research>}.
The frozen retriever returns \texttt{<information>}.
The observation is valid when at least one retrieved document is nonempty.
The accepted action enters \texttt{NEED\_EVIDENCE\_THINK}. \\
\texttt{NEED\_EVIDENCE\_}\newline\texttt{THINK} & Accept a substantive \texttt{<think>} that interprets the interpreter result.
A valid observation opens continuation or answering.
An invalid one returns to \texttt{NEED\_ACTION}. & Accept a substantive \texttt{<think>} that assesses the retrieved evidence.
A valid observation opens continuation or answering.
An invalid one returns to \texttt{NEED\_ACTION}. \\
\texttt{CAN\_CONTINUE\_}\newline\texttt{OR\_ANSWER} & Repeat \code{} and re-enter evidence reasoning, or accept a nonempty boxed answer and enter \texttt{DONE}. & Repeat retrieval with \texttt{<research>} and re-enter evidence reasoning.
Optionally accept one substantive \texttt{<summary>} as evidence organization, with no retriever call, no observation, and no phase change.
Alternatively, accept a nonempty \texttt{<answer>} and enter \texttt{DONE}. \\
\texttt{DONE} & Terminal.
Score the accepted answer by normalized numerical match. & Terminal.
Score the accepted answer by dataset-normalized exact match. \\
\bottomrule
\end{tabularx}
\end{table}

\FloatBarrier

\paragraph{Shift and Scale Invariance in GRPO.}
Let $\mathbf{R} = [R_1, \ldots, R_G]^\top$ denote the group return vector, $b$ a common additive shift, and $a > 0$ a positive scale factor.
Under the GRPO normalization in Equation~\ref{eq:grpo-advantage}, additive shifts are strictly invariant:
\begin{equation}
\label{eq:advantage-shift}
\widehat A_i(\mathbf{R} + b) = \widehat A_i(\mathbf{R}).
\end{equation}
For positive scaling $a > 0$, the normalized advantage transforms as:
\begin{equation}
\label{eq:advantage-scale}
\widehat A_i(a\mathbf{R}) = \frac{a(R_i - \bar R)}{a s_R + \epsilon} = \frac{R_i - \bar R}{s_R + \epsilon/a}, \qquad a > 0.
\end{equation}
These identities confirm that mean-centering eliminates additive offsets, while standard deviation scales linearly by $a$.
In the absence of the numerical stabilizer $\epsilon \to 0^+$, exact scale invariance holds whenever $s_R > 0$.
With a non-zero stabilizer $\epsilon > 0$, Equation~\ref{eq:advantage-scale} precisely characterizes the advantage adjustment when milestone rewards scale the return distribution.

\paragraph{When a group contains distinguishable returns.}
Consider a task prompt and a fixed policy generating independent, identically distributed trajectories.
Assume a sampled group contains $G \ge 2$ trajectories with exactly two possible distinct returns, where each trajectory achieves the higher return with probability $p \in (0, 1)$.
The empirical group variance $s_R^2$ is strictly positive if and only if both return values are present in the group:
\begin{equation}
\label{eq:mixed-return-probability}
\Pr(s_R > 0) = 1 - p^G - (1 - p)^G, \qquad G \ge 2.
\end{equation}
This holds because zero variance occurs exclusively in all-high ($p^G$) or all-low ($(1-p)^G$) groups.
The probability of non-zero within-group variance reaches its maximum at an equal mixture ($p = 0.5$) and vanishes as $p \to 0$ or $p \to 1$.
Consequently, increasing the success rate does not monotonically increase within-group return variation.
This two-return model illustrates a sampling constraint on gradient existence under sparse rewards rather than full multi-stage training dynamics.

\paragraph{From return differences to gradients.}
Now consider an unclipped expected-return objective under a fixed contract specification, task adapter, and environment.
Assuming a differentiable policy $\pi_\theta$ with parameter-independent support and sufficient regularity to swap differentiation and integration, the score-function gradient identity yields:
\begin{equation}
\label{eq:expected-return-score}
\nabla_\theta \mathbb E_{\tau \sim P_\theta}[R(\tau)] = \mathbb E_{\tau \sim P_\theta}\!\left[ (R(\tau) - \mathbb E[R(\tau)]) \nabla_\theta \log P_\theta(\tau) \right].
\end{equation}
Because the score function has zero expectation ($\mathbb E[\nabla_\theta \log P_\theta(\tau)] = 0$), subtracting the mean return leaves the overall gradient unchanged.
Since environment state transition dynamics are parameter-independent, only model-generated action tokens contribute directly to the trajectory score:
\begin{equation}
\label{eq:trajectory-score}
\nabla_\theta \log P_\theta(\tau) = \sum_t \nabla_\theta \log \pi_\theta(a_t \mid x_t^j).
\end{equation}
Non-zero return variance alone is insufficient to guarantee non-zero policy gradients; the trajectory return must also exhibit non-zero covariance with the trajectory score $\nabla_\theta \log P_\theta(\tau)$.
Loss-masked feedback modifies subsequent prompt contexts $x_t^j$ and action generation distributions without receiving direct gradient updates.
Note that Equation~\ref{eq:expected-return-score} serves as an idealized theoretical baseline; the practical implementation differs via GRPO group normalization, token-level averaging, PPO clipping, and KL regularization.

\paragraph{Design rationale.}
Initial milestone completions increment the recovery-adjusted progress credit $p_i$ and provide return variance when rollouts accumulate different recovery-adjusted totals.
Following each non-terminal transition $S_t^j \to S_{t+1}^j$, contract-state feedback $f_{t+1}$ updates the context $x_{t+1}^j$ for the subsequent decision step.
While feedback text carries no direct reward weight, action rejection reduces future verified progress credit via the recovery coefficient $\rho$.
When contract-state feedback is enabled, accepted non-terminal transitions receive positive phase guidance, whereas rejected transitions trigger recovery feedback and set $h_{t+1}=1$, so milestones first completed on the next accepted transition are discounted by $\rho$.

Deriving both feedback $f_{t+1}$ and credit $p_i$ from the unified contract verifier guarantees state-consistency between requested next actions and rewarded progress.
This structural coupling ensures implementation consistency without asserting equal necessity of feedback and credit.
This motivation specifically addresses the normalized advantage $\widehat{A}_i$ in GRPO; feedback modifies prompt contexts rather than directly modifying advantage estimates.
This derivation characterizes learning signal availability without claiming convergence bounds or guaranteed policy improvement.

\section{Skill Contracts and Visible Skill Instructions}
\label{app:skill-contracts}

\subsection{Exact Visible Skill Interface}
\label{app:exact-prompts-skills}

In the canonical skill-loading conditions, the initial context contains a system scaffold and a user message.\ %
The scaffold exposes the skill name, purpose, and loading rule, but not the visible skill instructions.\ %
After an accepted skill call, the runtime appends the visible skill instructions inside \texttt{<skill\_body>} and adds contract-state feedback with loss mask zero.\ %
Low-saturation purple marks skill content, blue marks policy output, green marks environment observations, orange marks feedback, and teal marks verified progress credit throughout the Appendix.\ %

\begin{lstlisting}[style=appendixcode]
Initial model context:
  [SYSTEM] skill catalog + generic loading rule
  [USER] question + skill-selection instruction

After an accepted skill call:
  [ASSISTANT] <skill_call>...</skill_call>
  [RUNTIME] <skill_body>...</skill_body>
            <hint>accepted state and next action</hint>

Each later transition:
  [ASSISTANT] one complete action
  [RUNTIME] optional environment observation
            optional <hint>...</hint>
\end{lstlisting}

\paragraph{Shared system scaffold.}
The shared scaffold uses the following two-entry catalog and loading rule:
\begin{lstlisting}[style=skillcode]
Part 1: Available skills
- name: code-interpreter-protocol
  purpose: Use this skill for mathematical problems that need reliable calculation, enumeration, symbolic checking, or program-aided verification.
- name: qa-search-protocol
  purpose: Use this skill for factual questions that need external evidence or multi-hop verification.

Part 2: Skill loading
Choose the useful skill before reasoning or answering. Output exactly <skill_call>SKILL_NAME</skill_call> and replace SKILL_NAME with one listed name. The harness will return its instructions inside <skill_body>...</skill_body>. Follow the returned instructions and environment observations.
\end{lstlisting}

Each single-task condition exposes only its corresponding catalog entry, while the Joint condition exposes both entries.\ %
These conditions use the following user message, with the preloaded Prompt exception described in Appendix~\ref{app:experimental-details}:
\begin{lstlisting}[style=appendixcode]
Question: {normalized question}

Select and load the useful skill before reasoning or answering.
\end{lstlisting}

\noindent\textbf{Math skill instructions.}\par
\begin{lstlisting}[style=skillcode]
Use Python calculations as evidence for the mathematical solution. First write one complete <think>...</think> to plan the calculation. Then put executable Python inside <code>...</code>. The code must perform the calculation and print a result. Wait for the environment-owned <interpreter>...</interpreter> observation and never write that tag yourself. Write another complete <think>...</think> to interpret the result. Repeat <code>...</code> and result reasoning when more calculation is useful. Finish with Answer: \boxed{...} only after analyzing a valid result.
\end{lstlisting}
Program-aided methods motivate executable computation~\bluecite{gao2023pal,chen2023programthoughts}.\ %
Tool-integrated methods motivate repeated action-observation reasoning~\bluecite{gou2024tora,deepseekai2025retool}.\ %

\noindent\textbf{Search skill instructions.}\par
\begin{lstlisting}[style=skillcode]
Use retrieved evidence to answer the factual question. First write one complete <think>...</think> to plan the retrieval. Use <search>...</search> for the first concise query. Wait for the environment-owned <information>...</information> observation and never write that tag yourself. Write another complete <think>...</think> to assess the evidence. If more evidence is needed, use a refined <research>...</research> query and repeat result reasoning. An optional <summary>...</summary> may organize gathered evidence but does not replace reasoning. Finish with <answer>...</answer> only after analyzing valid evidence.
\end{lstlisting}
The retrieval loop follows Search-R1~\bluecite{jin2025searchr1}.\ %
AutoRefine~\bluecite{shi2025autorefine} and ReSum~\bluecite{wu2025resum} are adjacent precedents for evidence refinement and context summarization rather than reproduced algorithms.\ %

\section{Experimental Details and Provenance}
\label{app:experimental-details}

\subsection{Training Configuration}

All online experiments use the open-source \texttt{slime} framework with its Megatron optimization backend and SGLang rollout engine~\bluecite{slime_github}.\ %
The reported runs use four GPUs and tensor parallelism two.\ %
Math and Search use 200 rollout rounds, while all Joint conditions, including the recovery comparison, use 400.\ %
Math training contains 7,500 MATH and 7,473 GSM8K examples.\ %
Search training uses 79,168 Natural Questions and 90,447 HotpotQA source rows, and Joint combines all 14,973 Math examples with 13,827 sampled Search examples.\ %

Training uses Adam with learning rate $10^{-6}$, betas $(0.9,0.98)$, warmup over 0.285 of the planned optimizer updates, weight decay 0.01, entropy coefficient 0.001, KL coefficient 0.003, gradient clipping 1.0, policy clipping $(0.20,0.28)$, and TIS clipping 4.0.\ %
Rollout and evaluation temperatures are 1.\ %
Table~\ref{tab:training_setup} gives the task-specific batch and update budgets.\ %
Contract training and evaluation allow at most 20 executor turns per trajectory, including rejected actions, and 8 environment calls.\ %
Each executor turn generates at most 1,024 tokens, subject to a total context limit of 16,384 tokens that includes the prompt, generated actions, observations, and contract-state feedback.\ %
Exhausting the turn or token budget without accepted termination counts as protocol failure.\ %

The KL reference policy $\pi_{\mathrm{ref}}$ is frozen at the executor's initial checkpoint, distinct from the sampling policy $\pi_{\theta_{\mathrm{old}}}$.\ %
For each trainable token, the regularizer in Equation~\ref{eq:grpo-objective} uses the sampled approximation
\[
\delta_{i,\ell}=\log\frac{\pi_\theta(a_{i,\ell}\mid x_{i,\ell})}{\pi_{\mathrm{ref}}(a_{i,\ell}\mid x_{i,\ell})},
\qquad
d^{\mathrm{KL}}_{i,\ell}=\operatorname{clip}\!\left(e^{-\delta_{i,\ell}}-1+\delta_{i,\ell},-10,10\right).
\]

The 500-example monitor runs every 25 rounds using a separate evaluation plan, not necessarily the training datasets.\ %
For Figure~\ref{fig:acquisition-dynamics}, Math uses 250 MATH and 250 GSM8K examples, while Search uses 250 Natural Questions and 250 MuSiQue examples.\ %
Figure~\ref{fig:acquisition-dynamics} compares matched Outcome-only and Credit-only conditions with feedback disabled.\ %
All-zero groups remain in training and receive zero normalized advantages after group centering.\ %
Each plotted point is one monitor evaluation, so no repeated-run interval is reported.\ %
Search uses an E5-base-v2 retriever over frozen 2018 Wikipedia with a FAISS index and top-$3$ retrieval~\bluecite{jin2025searchr1}.\ %
Math uses a restricted Python sandbox with nonempty output and fixed resource limits.\ %

\begin{table}[H]
\caption{Online training configurations.\ %
P/R is prompts per rollout, S/P is samples per prompt, and U/R is optimizer updates per rollout.\ %
Model-scale runs reuse the corresponding task profile, and the Feedback + Credit conditions in RQ2 reuse the main single-task runs.}
\label{tab:training_setup}
\centering
\small
\setlength{\tabcolsep}{3.4pt}
\renewcommand{\arraystretch}{1.07}
\begin{tabular}{@{}lrrrrrrr@{}}
\toprule
Condition & GPU & TP & Rounds & P/R & S/P & U/R & Batch \\
\midrule
O-Math / Math interventions & 4 & 2 & 200 & 72 & 5 & 5 & 72 \\
O-Search / Search interventions & 4 & 2 & 200 & 512 & 5 & 10 & 256 \\
O-Joint & 4 & 2 & 400 & 72 & 5 & 5 & 72 \\
Recovery-coefficient comparison & 4 & 2 & 400 & 72 & 5 & 5 & 72 \\
\bottomrule
\end{tabular}
\end{table}

\FloatBarrier

\subsection{Evaluation and Baselines}

Each full-suite condition is evaluated on 6,319 Math and 51,713 Search examples.\ %
The Math suite contains 5,000 MATH and 1,319 GSM8K examples.\ %
The Search suite contains 3,610 Natural Questions, 11,313 TriviaQA, 14,267 PopQA, 7,405 HotpotQA, 12,576 2WikiMultiHopQA, 2,417 MuSiQue, and 125 Bamboogle examples.\ %
Direct omits the contract runtime, while Prompt preloads the skill and uses the contract runtime without training for all executors.\ %
Both use one sample per problem and serve as deployment context rather than controls for Outcome-only RL.\ %

These Direct and Prompt runs score an answer extracted from the whole response, even when the runtime has not accepted termination.\ %
Their differences measure behavior under different evaluation conditions, not a loss of model knowledge.\ %

For skill-contract evaluation, Equations~\ref{eq:pcr} and~\ref{eq:rf-pcr} define PCR and RF-PCR from accepted termination and the recovery flag.\ %
RF-PCR further requires that no rejected-action feedback occur under the runtime with feedback enabled.\ %
It can include feedback after accepted actions and is distinct from disabling feedback in RQ3.\ %
For contract-trained policies, Task Outcome scores the accepted final answer, while Hint Rate is the percentage of samples that receive at least one rejected-action feedback message.\ %
We do not pool Math and Search or combine Protocol Completion with Task Outcome.\ %

The ReTool-style Math row is a matched adaptation that shares the sandbox, data, rollout budget, and pass@1 task verifier with O-Math.\ %
Its incorrect-trajectory score is capped at $-0.6$, while the referenced recipe uses $0$.\ %
The Search-R1 Recipe row shares the prepared data, retriever, corpus, action environment, and evaluator with O-Search.\ %
These matched rows provide local performance context rather than untouched official reproductions.\ %

\section{Complete Model-Visible Execution Traces}
\label{app:execution-traces}

The following correct examples come from the Ours-S policies.\ %
Role labels and LaTeX-safe punctuation replace chat-template formatting, while message wording and environment observations are otherwise retained.\ %

\subsection{Math}

\begin{trajectorybox}{Math | O-Math | Correct and observation-grounded}

\begin{tracebox}[systembg]{System}
Part 1: Available skills\\
\texttt{- name: code-interpreter-protocol}\\
purpose: Use this skill for mathematical problems that need reliable calculation, enumeration, symbolic checking, or program-aided verification.\\[2pt]
Part 2: Skill loading\\
Choose the useful skill before reasoning or answering. Output exactly \texttt{<skill\_call>SKILL\_NAME</skill\_call>} and replace \texttt{SKILL\_NAME} with one listed name. The harness will return its instructions inside \texttt{<skill\_body>...</skill\_body>}. Follow the returned instructions and environment observations.\ %
\end{tracebox}

\begin{tracebox}[userbg]{User}
Question: A circle of radius 5 with its center at $(0,0)$ is drawn on a Cartesian coordinate system. How many lattice points (points with integer coordinates) lie within or on this circle?\\[2pt]
Select and load the useful skill before reasoning or answering.\ %
\end{tracebox}

\begin{tracebox}[policybg]{Assistant}
\texttt{<skill\_call>code-interpreter-protocol</skill\_call>}
\end{tracebox}

\begin{tracebox}[skillbg]{Runtime skill body}
\texttt{<skill\_body>}Use Python calculations as evidence for the mathematical solution. First write one complete \texttt{<think>...</think>} to plan the calculation. Then put executable Python inside \texttt{<code>...</code>}. The code must perform the calculation and print a result. Wait for the environment-owned \texttt{<interpreter>...</interpreter>} observation and never write that tag yourself. Write another complete \texttt{<think>...</think>} to interpret the result. Repeat \texttt{<code>...</code>} and result reasoning when more calculation is useful. Finish with $\text{Answer: }\boxed{\ldots}$ only after analyzing a valid result.\texttt{</skill\_body>}\\[2pt]
\end{tracebox}

\begin{tracebox}[feedbackbg]{Contract-state feedback}
\texttt{<hint>Action accepted. State: the skill is loaded. Next: write one complete <think>...</think> plan.</hint>}
\end{tracebox}

\begin{tracebox}[policybg]{Assistant}
\texttt{<think>}We need to count all lattice points within and on the circle of radius 5 centred at the origin $(0,0)$. The equation of the circle is $x^2+y^2\leq25$. We need to check all integer pairs $(x,y)$ satisfying this condition.\texttt{</think>}
\end{tracebox}

\begin{tracebox}[feedbackbg]{Contract-state feedback}
\texttt{<hint>Action accepted. State: Python evidence is required. Next: run calculation in <code>...</code> and print a result.</hint>}
\end{tracebox}

\begin{tracelisting}[policybg]{Assistant code action}
<code>
count = 0
for x in range(-5, 6):
    for y in range(-5, 6):
        if x**2 + y**2 <= 25:
            count += 1
print(count)
</code>
\end{tracelisting}

\begin{tracebox}[observationbg]{Environment observation}
\texttt{<interpreter>Output: 81</interpreter>}
\end{tracebox}

\begin{tracebox}[feedbackbg]{Contract-state feedback}
\texttt{<hint>Action accepted. State: an environment observation is ready. Next: write one complete <think>...</think>.</hint>}
\end{tracebox}

\begin{tracebox}[policybg]{Assistant}
\texttt{<think>}Based on the written Python script, the number of lattice points within or on the circle is 81.\texttt{</think>}
\end{tracebox}

\begin{tracebox}[feedbackbg]{Contract-state feedback}
\texttt{<hint>Action accepted. State: evidence is analyzed. Next: use <code>...</code> again or give Answer: $\boxed{\ldots}$.</hint>}
\end{tracebox}

\begin{tracebox}[creditbg]{Final answer}
$\text{Answer: }\boxed{81}$
\end{tracebox}

\end{trajectorybox}

\subsection{Search}

\begin{trajectorybox}{Search | O-Search | Correct and observation-grounded with evidence organization}

\begin{tracebox}[systembg]{System}
Part 1: Available skills\\
\texttt{- name: qa-search-protocol}\\
purpose: Use this skill for factual questions that need external evidence or multi-hop verification.\\[2pt]
Part 2: Skill loading\\
Choose the useful skill before reasoning or answering. Output exactly \texttt{<skill\_call>SKILL\_NAME</skill\_call>} and replace \texttt{SKILL\_NAME} with one listed name. The harness will return its instructions inside \texttt{<skill\_body>...</skill\_body>}. Follow the returned instructions and environment observations.\ %
\end{tracebox}

\begin{tracebox}[userbg]{User}
Question: who played harmonica on bring it on home?\\[2pt]
Select and load the useful skill before reasoning or answering.\ %
\end{tracebox}

\begin{tracebox}[policybg]{Assistant}
\texttt{<skill\_call>qa-search-protocol</skill\_call>}
\end{tracebox}

\begin{tracebox}[skillbg]{Runtime skill body}
\texttt{<skill\_body>}Use retrieved evidence to answer the factual question. First write one complete \texttt{<think>...</think>} to plan the retrieval. Use \texttt{<search>...</search>} for the first concise query. Wait for the environment-owned \texttt{<information>...</information>} observation and never write that tag yourself. Write another complete \texttt{<think>...</think>} to assess the evidence. If more evidence is needed, use a refined \texttt{<research>...</research>} query and repeat result reasoning. An optional \texttt{<summary>...</summary>} may organize gathered evidence but does not replace reasoning. Finish with \texttt{<answer>...</answer>} only after analyzing valid evidence.\texttt{</skill\_body>}\\[2pt]
\end{tracebox}

\begin{tracebox}[feedbackbg]{Contract-state feedback}
\texttt{<hint>Action accepted. State: the skill is loaded. Next: write one complete <think>...</think> plan.</hint>}
\end{tracebox}

\begin{tracebox}[policybg]{Assistant}
\texttt{<think>}I will begin by running a query focusing on the musician who played harmonica in the song ``Bring It On Home''. My first query should be concise and relevant.\texttt{</think>}
\end{tracebox}

\begin{tracebox}[feedbackbg]{Contract-state feedback}
\texttt{<hint>Action accepted. State: retrieval evidence is required. Next: use a concise <search>...</search> query.</hint>}
\end{tracebox}

\begin{tracebox}[policybg]{Assistant}
\texttt{<search>who played harmonica on bring it on home?</search>}
\end{tracebox}

\begin{tracelisting}[observationbg]{Environment observation}
<information>
Doc 1: "Bring It On Home (Sonny Boy Williamson II song)"
Bring It On Home (Sonny Boy Williamson II song) ""Bring It On Home"" is a blues song written by American music arranger and songwriter Willie Dixon. The first known recording of the song was by Sonny Boy Williamson II in 1963. It was later recorded by Led Zeppelin in 1969, and has been covered by many artists since. Sonny Boy Williamson's version of the song was recorded on January 11, 1963 in Chicago. Accompanying Williamson on vocals and harmonica were Matt ""Guitar"" Murphy on guitar, Milton Rector on bass guitar, Al Duncan on drums, and either Lafayette Leake or Billy

\end{tracelisting}

\begin{tracelisting}[observationbg]{Environment observation (continued)}
Doc 2: "Bring It On Home (album)"
Bring It On Home (album) Bring It On Home is the seventh studio album by Joan Osborne released under Saguaro Road Records on March 27, 2012. It was her first album in five years. On this album she co-produced with the guitarist Jack Petruzzelli. The album consists entirely of Blues and R&B covers. The record also includes tracks originally made famous by American blues masters, such as Sonny Boy Williamson (""Bring it on Home""), Muddy Waters (""I Want to Be Loved""), as well as recordings originally released by some of the best-known R&B performars---including Ray Charles (""I Don't Need No

\end{tracelisting}

\begin{tracelisting}[observationbg]{Environment observation (continued)}
Doc 3: "Bring It On Home (Sonny Boy Williamson II song)"
in the concert film ""The Song Remains the Same"".
\end{tracelisting}

\begin{tracelisting}[observationbg]{Environment observation (continued)}
""Bring It On Home"" was played at the reunion of surviving Led Zeppelin members staged at Jason Bonham's wedding reception in May 1990. Bring It On Home (Sonny Boy Williamson II song) ""Bring It On Home"" is a blues song written by American music arranger and songwriter Willie Dixon. The first known recording of the song was by Sonny Boy Williamson II in 1963. It was later recorded by Led Zeppelin in 1969, and has been covered by many artists since. Sonny Boy Williamson's version of the song was recorded on
</information>
\end{tracelisting}

\begin{tracebox}[feedbackbg]{Contract-state feedback}
\texttt{<hint>Action accepted. State: an environment observation is ready. Next: write one complete <think>...</think>.</hint>}
\end{tracebox}

\begin{tracebox}[policybg]{Assistant}
\texttt{<think>}The document titled ``Bring It On Home (Sonny Boy Williamson II song)'', provides evidence that Sonny Boy Williamson played harmonica on the song ``Bring It On Home''.\texttt{</think>}
\end{tracebox}

\begin{tracebox}[feedbackbg]{Contract-state feedback}
\texttt{<hint>Action accepted. State: evidence is analyzed. Next: use <research>...</research> again, organize it with <summary>...</summary>, or give <answer>...</answer>.</hint>}
\end{tracebox}

\begin{tracebox}[policybg]{Assistant}
\texttt{<summary>}Based on the information from the music document on the 1963 recording of Sonny Boy Williamson's ``Bring It On Home,'' the harmonica player for this song was indeed Sonny Boy Williamson. \texttt{</summary>}
\end{tracebox}

\begin{tracebox}[feedbackbg]{Contract-state feedback}
\texttt{<hint>Action accepted. State: evidence is summarized. Next: use <research>...</research> again or give <answer>...</answer>.</hint>}
\end{tracebox}

\begin{tracebox}[creditbg]{Final answer}
\texttt{<answer>Sonny Boy Williamson</answer>}
\end{tracebox}

\end{trajectorybox}

\section{Controlled Intervention Results}
\label{app:controlled-results}

RQ2 compares independently trained Qwen2.5-3B conditions for training signals and the recovery coefficient.\ %
RQ3 tests inference-time interventions on frozen O-Math and O-Search checkpoints.\ %
PCR denotes completion under the runtime configured for each condition and can include recovery after rejected-action feedback.\ %
Hint Rate is the percentage of samples that receive at least one rejected-action feedback message.\ %

The feedback-credit factorial, recovery comparison, and visible skill interventions each use MATH and GSM8K for Math and 2WikiMultiHopQA and HotpotQA for Search.\ %
We interpret each PCR value only within its intervention and do not pool these controls as one evaluation pass.\ %

\subsection{RQ2: Training Behavior}

\subsubsection{Feedback-Credit Factorial}

Each condition starts from the same base checkpoint within its task.\ %
All four retain the same task reward, training data, rollout budget, contract verifier, and visible skill instructions.\ %
Outcome-only removes both contract-state feedback and verified progress credit, Feedback-only retains feedback, Credit-only retains credit, and Feedback + Credit retains both.\ %
In Outcome-only, a rejected action receives no negative reward, preserves the phase, and does not terminate the trajectory.\ %
Only a final answer accepted after all required phases is sent to the binary task verifier.\ %
The factorial therefore tests learning under one contract runtime rather than comparing Feedback + Credit with permissive terminal-only RL.\ %
It does not assume a symmetric interaction between the two components.\ %
Figure~\ref{fig:acquisition-dynamics} reports the complete monitor curves for the matched Outcome-only and Credit-only pair.\ %

\begin{table}[H]
\caption{Training-time feedback-credit factorial on Qwen2.5-3B.\ %
Each condition is trained independently from the same base checkpoint within its domain.\ %
The task reward and contract runtime are fixed, while Feedback toggles contract-state feedback and Credit toggles verified progress credit.\ %
Outcome-only uses the contract runtime, in which rejected actions do not receive a penalty or terminate the trajectory, but only a final answer accepted after all required phases can receive the task reward.\ %
Task Outcome and PCR are unweighted averages over MATH and GSM8K for Math and 2WikiMultiHopQA and HotpotQA for Search.}
\label{tab:codesign_ablation}
\centering
\small
\setlength{\tabcolsep}{5pt}
\renewcommand{\arraystretch}{1.08}
\begin{tabular}{llrrrr}
\toprule
Domain & Condition & Feedback & Credit & Outcome & PCR \\
\midrule
Search & B00 Outcome-only & 0 & 0 & 0.00 & 0.00 \\
Search & B10 Feedback-only & 1 & 0 & 0.00 & 0.00 \\
Search & B01 Credit-only & 0 & 1 & 29.28 & 98.69 \\
Search & B11 Feedback + Credit & 1 & 1 & 30.11 & 98.90 \\
\addlinespace
Math & B00 Outcome-only & 0 & 0 & 0.00 & 0.00 \\
Math & B10 Feedback-only & 1 & 0 & 0.00 & 0.00 \\
Math & B01 Credit-only & 0 & 1 & 59.76 & 93.17 \\
Math & B11 Feedback + Credit & 1 & 1 & 64.61 & 97.29 \\
\bottomrule
\end{tabular}
\end{table}

\FloatBarrier

\subsubsection{Recovery Coefficient}

The main setting uses $\rho=0.90$, while the no-discount condition uses $\rho=1.00$.\ %
Both are independently trained Ours-J policies with the same task mixture and runtime.\ %
Main-text Table~\ref{tab:runtime-recovery-boundary} reports the result.\ %
For Math with $\rho=1.00$, RF-PCR is zero despite high PCR, indicating that every trajectory reaching accepted termination receives rejected-action feedback.\ %
RF-PCR still permits feedback after accepted actions and does not measure execution without feedback.\ %

\subsection{RQ3: Inference Behavior}

\subsubsection{Visible Skill Interventions}
\label{app:skill-interventions}

S0 uses the original visible skill instructions, S1 removes them, S2 uses the independently written paraphrases below, and S3 injects the visible skill instructions of the other task.\ %
Only the visible skill instructions $z_j$ change in this test.\ %
The contract runtime and task verifier remain fixed.\ %
The result therefore identifies an effect of the visible skill instructions while the rest of the skill contract remains fixed.\ %
S3 changes the action grammar and is therefore a compatibility stress test rather than a grammar-controlled semantic contradiction.\ %

\paragraph{S2 paraphrased instructions.}
\label{app:paraphrased-skills}
The registered S2 intervention replaces $z_j$ with an independently written paraphrase.\ %
The Math rewrite changes the opening instruction, describes the plan as identifying the needed calculation, and expresses repeated execution as alternating code and reasoning.\ %
The Search rewrite describes the plan as identifying what to retrieve, evaluates relevance after each observation, and requests a revised query when evidence is insufficient.\ %
Both rewrites retain the action inventory and order, environment ownership of observations, repeated-action permission, and final-answer format.\ %

\noindent\textbf{Paraphrased Math skill instructions.}\par
\begin{lstlisting}[style=skillcode]
Solve the mathematical problem with evidence produced by Python. Begin with one complete <think>...</think> that identifies the needed calculation. Place executable Python in <code>...</code> and print the computed value. Only the environment may return <interpreter>...</interpreter>. After each result, write a complete <think>...</think> that interprets it. Alternate further code and reasoning when needed. Give Answer: \boxed{...} after using a valid result.
\end{lstlisting}

\noindent\textbf{Paraphrased Search skill instructions.}\par
\begin{lstlisting}[style=skillcode]
Answer the factual question from retrieved evidence. Start with one complete <think>...</think> that identifies what to retrieve. Put the first focused query in <search>...</search>. Only the environment may return <information>...</information>. After each observation, write a complete <think>...</think> that evaluates its relevance. Use a revised <research>...</research> query when evidence is insufficient. A <summary>...</summary> may organize evidence. Finish with <answer>...</answer> only after evidence analysis.
\end{lstlisting}

\begin{table}[H]
\caption{Inference-time visible skill instruction interventions on frozen Qwen2.5-3B single-task checkpoints.\ %
Only the visible skill instructions $z_j$ change.\ %
Paraphrased rewrites wording, plan descriptions, and transition language while preserving action tags, action order, environment-owned observations, repeated-action permission, and the termination format.\ %
Appendix~\ref{app:paraphrased-skills} gives the exact Math and Search bodies.\ %
Outcome is pass@1 for Math and exact match for Search.}
\label{tab:skill_counterfactual}
\centering
\footnotesize
\setlength{\tabcolsep}{2.2pt}
\begin{tabular}{llrrrrrrr}
\toprule
Domain & Visible skill & MATH & GSM & 2Wiki & Hotpot & Avg. & PCR & Hint Rate \\
\midrule
Math & Original & 48.62 & 80.59 & -- & -- & 64.61 & 97.29 & 5.59 \\
Math & Removed & 42.76 & 74.68 & -- & -- & 58.72 & 91.34 & 11.70 \\
Math & Paraphrased & 43.92 & 75.59 & -- & -- & 59.75 & 94.36 & 7.27 \\
Math & Wrong Search & 1.48 & 2.35 & -- & -- & 1.92 & 4.71 & 99.69 \\
\addlinespace
Search & Original & -- & -- & 29.87 & 30.34 & 30.11 & 98.90 & 0.99 \\
Search & Removed & -- & -- & 22.57 & 18.51 & 20.54 & 63.81 & 14.86 \\
Search & Paraphrased & -- & -- & 26.92 & 26.49 & 26.71 & 98.86 & 1.08 \\
Search & Wrong Math & -- & -- & 29.00 & 30.93 & 29.96 & 98.31 & 1.74 \\
\bottomrule
\end{tabular}
\end{table}

\FloatBarrier

\subsubsection{Observation and Feedback Interventions}

For observation withholding, the environment first executes the admissible action and the task adapter validates the returned observation.\ %
The runtime then replaces only the model-visible payload with a non-evidential placeholder while preserving the recorded validity and contract transition.\ %
This intervention removes observation content available to the executor without changing whether the original action and observation passed runtime validation.\ %
The four dataset-level changes are reported in Section~\ref{sec:inference-behavior} and visualized in Figure~\ref{fig:contract-acquisition}(c).\ %

The feedback-removal intervention suppresses the complete feedback channel at inference while retaining validation.\ %
Its result is reported in main-text Table~\ref{tab:runtime-recovery-boundary}.\ %

\section{Discussion}
\label{app:discussion}

Grounded skill-following treats a fixed expert-authored skill as a multi-step behavioral standard rather than a local prompt.\ %
A skill contract makes this behavior trainable and auditable by verifying ordered progress.\ %
Under the tested contract runtime, Credit-only develops nonzero PCR and Task Outcome while Outcome-only remains at zero.\ %
PCR and Task Outcome report Protocol Completion and answer correctness separately, while RF-PCR identifies completion without rejected-action feedback.\ %

The Math/Search interventions and joint/separate comparisons suggest that high Protocol Completion under a common contract design can coexist with task-dependent reliance on execution signals and differing Task Outcome gains.\ %
Motivated by prior studies of skill neurons and layer-aware task arithmetic~\bluecite{wang2022skillneurons,chen2025layeraware}, we hypothesize that these capabilities involve both shared and task-specific parameters or representations.\ %
Our behavioral results do not identify such components, and differences in task exposure further limit the interpretation of joint versus separate training.\ %
Testing this hypothesis would require targeted interventions on parameters or representations under controlled training exposure.\ %

The skill contracts are manually specified.\ %
The interventions establish dependence on visible skill instructions, observations, and contract-state feedback within this implementation, but not token-level causal evidence use or scaffold-free internalization.\ %
The results establish the proposed problem and method in the tested Math, Search, and five Qwen executor configurations.\ %
They do not establish automatic contract construction, universal transfer, guaranteed optimization improvement, or failure of terminal-only RL under a permissive runtime.\ %

\end{document}